%% file: neurips_2026.tex
\documentclass[letterpaper]{article} 
\usepackage{aaai2026}  
\usepackage{times}  
\usepackage{helvet}  
\usepackage{courier}  
\usepackage[hyphens]{url}  
\usepackage{graphicx} 
\usepackage{natbib}  
\usepackage{caption} 
\usepackage{algorithm}
\usepackage{algorithmic}
\usepackage{subfigure}
\usepackage{tabularray}
\usepackage{multirow}
\usepackage{graphicx}      
\usepackage{booktabs} 
\usepackage[table]{xcolor}
\usepackage{amsmath}
\usepackage{amssymb}
\usepackage{newfloat}
\usepackage{listings}
\DeclareCaptionStyle{ruled}{labelfont=normalfont,labelsep=colon,strut=off} 
\floatstyle{ruled}
\newfloat{listing}{tb}{lst}{}
\floatname{listing}{Listing}
\title{Entropy-Gradient Inversion: Moving Toward Internal Mechanism of Large  Reasoning Models}

\author{%
Junyao Yang$^{1}$\thanks{Equal contribution.} , 
  Chen Qian$^{2}$\footnotemark[1] ,
  Kun Wang$^{4}$, 
  Linfeng Zhang$^{3}$, \\
  \textbf{Quanshi Zhang$^{3}$, Yong Liu$^{2}$\thanks{Corresponding author.}, Dongrui Liu$^{3}$\footnotemark[2]} \\
    $^1$National University of Singapore, $^2$Renmin University of China \\
  $^3$Shanghai Jiao Tong University,
  $^4$Nanyang Technological University \\
  \texttt{junyao.yang@u.nus.edu}
}
  
\makeatletter
\providecommand{\copyright@text}{}
\makeatother
\begin{document}

\maketitle

\input{abstract}

\input{sections/introduction}
\input{sections/entropy_gradient_inversion}

\input{sections/method}

\input{sections/main_experiment}

\input{sections/related_work}

\input{sections/conclusions}


\clearpage
\bibliography{references}

\clearpage
\addtocontents{toc}{\protect\setcounter{tocdepth}{1}} 

\input{appendix}


\newpage

\input{checklist.tex}
\end{document}

%% file: abstract.tex
\begin{abstract}
The advancement of Large Reasoning Models (LRMs) has catalyzed a paradigm shift from reactive ``fast thinking'' text generation to systematic, step-by-step ``slow thinking'' reasoning, unlocking state-of-the-art performance in complex mathematical and logical tasks. However, the field faces \textit{the fundamental gap between token-level behavioral analysis and internal reasoning mechanisms, and the instability of reinforcement learning (RL) for reasoning optimization relying on costly external verifiers}. We identify and formally define \textbf{Entropy-Gradient Inversion}, a robust negative correlation between token entropy and logit gradients that acts as a definitive geometric fingerprint for LRM reasoning capability. Building on this, we propose \textbf{Correlation-Regularized Group Policy Optimization (CorR-PO)}, which embeds this inversion signature into RL reward regularization. Extensive experiments on various reasoning benchmarks across multiple model scales show CorR-PO consistently outperforms state-of-the-art baselines, confirming that stronger inversion directly correlates with superior reasoning performance.
\end{abstract}

%% file: sections/introduction.tex
\section{Introduction}
The development of Large Reasoning Models (LRMs) shifts away from quick text generation and toward careful, step-by-step reasoning \citep{grattafiori2024llama,qwen2.5,yang2025qwen3technicalreport}. This paradigm shift, frequently analogized to the cognitive transition from ``fast thinking'' to ``slow thinking'' \citep{kahneman2011thinking,jaech2024openai,guo2025deepseek}, has unlocked unprecedented capabilities in complex domains such as mathematics, logic, and competitive programming \citep{openai2025o3,qwq32b}. 
Recent breakthroughs, notably DeepSeek-R1 and OpenAI o1-like models \citep{guo2025deepseek, jaech2024openai}, have demonstrated that these profound reasoning behaviors, such as self-verification, reflection, and strategic backtracking, can emerge purely through large-scale Reinforcement Learning (RL) \citep{guo2024deepseekmath}.

To reveal the mechanism of LRMs' reasoning capabilities, token-level entropy has emerged as a central diagnostic signal that quantifies step-wise {uncertainty} and governs the balance between exploration and exploitation in RL-driven reasoning, whereby high-entropy ``forking'' tokens drive policy improvement and entropy collapse precipitates premature convergence \citep{wangbeyond20258020rulehighentropyminority,cui2025entropymechanismreinforcementlearning,qian2025demystifyingreasoningdynamicsmutual}. 
Beyond optimization, entropy consistently serves as a critical metric for cognitive transitions in LRMs, including pivotal guiding tokens, {reflective} verification actions, and the emergence of {rare} alternative solutions throughout reasoning trajectories
\citep{li-etal-2025-happened,cheng2025reasoningexplorationentropyperspective,cui2025entropymechanismreinforcementlearning}. 
While prior work has established an inextricable link between token entropy and LLM reasoning, a natural question arises: why do reasoning models exhibit such an entropy feature? 
In addition, what underlying mechanisms connect entropy to the internal representations of the model? 
Although entropy provides valuable token-level uncertainty signals, {\textbf{fully elucidating the ``slow thinking'' requires moving beyond output sequences to investigate the underlying architecture and evolution of internal hidden states}} \citep{zhang2025truth, zhang2025tracing,saphra2024mechanistic, li2025reasoning}. 

To bridge this gap between external behavior and internal mechanisms, we leverage gradient-based analysis as a diagnostic tool that couples internal parameter sensitivities with output-level features \citep{li-etal-2025-happened}, thereby exposing geometric correlations that structurally characterize the reasoning process in LRMs \citep{ abnar2023transformer, yang2026reasonanyincorporatingreasoningcapability}.
Driven by this motivation, we conduct a preliminary exploration using state-of-the-art models to calculate the correlation between token-level prediction entropy and the $L_1$ norm of corresponding logit gradients.
As illustrated in Figure \ref{fig:intro_fig}, the bottom part shows that base models and safety-aligned models without RL training exhibit \textbf{\textit{No Significant Correlation}} between token-level entropy and logit gradients.
Notably, in the context of LRMs, we observe a striking phenomenon where \textit{\textbf{LRMs demonstrate a strong negative correlation between token-level entropy and logit gradients across all evaluation datasets}}, a behavior that we formally define as the \textbf{\textit{Entropy-Gradient Inversion}} phenomenon, as illustrated in the upper part of Figure \ref{fig:intro_fig} and detailed in Section \ref{subsec:entropy_gradient_inversion_subsec}2.2.
To further investigate how the inversion effect evolves across model training, we characterize the \textbf{\textit{Training Dynamics}} of the Entropy-Gradient phenomenon, as detailed in Section \ref{sec:training_dynamics}2.3. 
Surprisingly, we observe that this inversion effect emerges rapidly during Supervised Fine-Tuning (SFT) \citep{hu2021loralowrankadaptationlarge} and strengthens further via Reinforcement Learning (RL), an observation that tightly links this signature’s emergence to the model’s developing reasoning capability.
This negative correlation serves as a fingerprint for the mastery of slow thinking trajectories within LRMs compared to standard instruct-tuned models, progressively strengthening throughout training stages.

\input{multi_plots/intro_fig}


Driven by the signal of Entropy-Gradient Inversion to reveal the internal mechanism of LRMs, we observe that this signature tightly couples with strong reasoning capability yet only gradually realizes across post-training process,  wondering if it is possible that \textbf{the Entropy-Gradient Inversion pattern can be directly incorporated as an inductive prior into the reinforcement learning training process of LRMs},
 thereby accelerating the acquisition of stronger reasoning capability and achieving more stable state-of-the-art reasoning performance.
Based on this key insight, we propose \textbf{Cor}relation-\textbf{R}egularized Group \textbf{P}olicy \textbf{O}ptimization (\textbf{CorR-PO}). Building on Group Relative Policy Optimization (GRPO) \citep{guo2024deepseekmath}, CorR-PO embeds our discovered geometric heuristic directly into the reward function. Specifically, we compute the Spearman correlation coefficient \(\rho_{E,I}\) \citep{spearman1961proofspearman} between reasoning trajectory \textbf{Average Step Entropy} and logit gradients as textbf{Internal Gradient Influence}, and design a \textbf{Correlation Regularization Reward} \(R_{\text{corr}} = -(1+\rho_{E,I})\) to penalize weak or positive correlations linked to ``fast thinking''. 
This regularizes the model’s latent space to form reasoning structures early in RL training, and experiments show CorR-PO outperforms state-of-the-art baselines across benchmarks.

%% file: multi_plots/intro_fig.tex
\begin{figure*}[t]
    \centering
      \centering

    \subfigure{
        \includegraphics[width=0.8\textwidth]{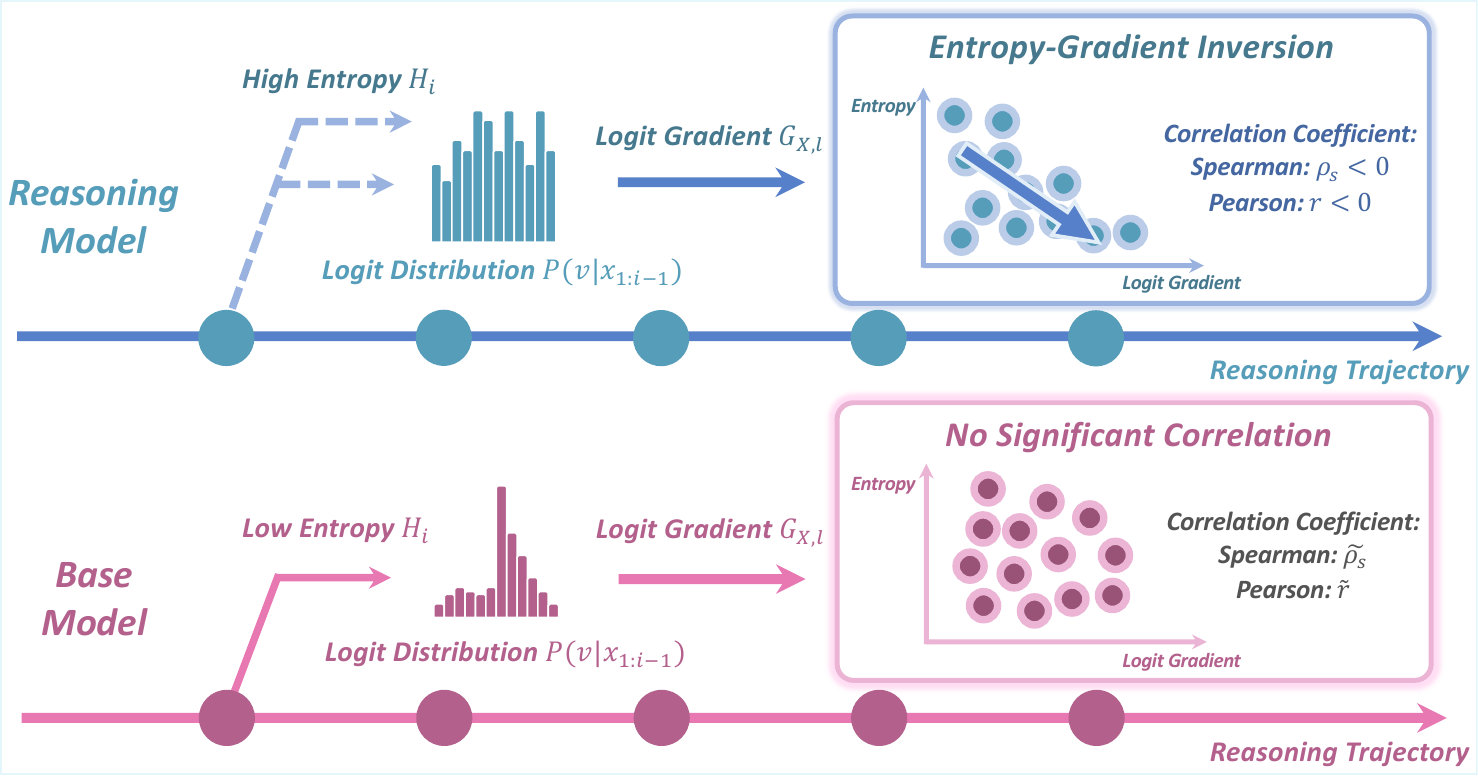}
        \label{fig:intro_fig}
    }
    
    \vspace{-0.5em}
    \caption{\textbf{Illustration of the Entropy-Gradient Inversion.} \textbf{Top part}: In \textbf{Base Models}, output entropy and logit gradients exhibit \textbf{No Significant Correlation}, shown as both spearman $\rho_s$ and pearson $r$ correlation coefficient . \textbf{Bottom part}: \textbf{Reasoning Models} demonstrate the \textbf{Entropy-Gradient Inversion}, serving as a fingerprint for the transition to the {slow thinking}.}
    \label{fig:intro_fig}
\end{figure*}

%% file: sections/entropy_gradient_inversion.tex
\section{Entropy-Gradient Inversion: An Internal Geometric Metric of LRMs' Slow Thinking Mechanism}

\label{sec:entropy_gradient_inversion_main}

To investigate the internal mechanism in LRMs \citep{li202512surveyreasoning,xu2025largereasoningmodelssurvey}, we examine how prediction entropy relates to internal gradients in this section.
Subsection 2.1\ref{subsec:preliminaries_entropy_gradient} formalizes token-level gradient influence, entropy, and their correlation.
Subsection 2.2\ref{subsec:entropy_gradient_inversion_subsec} compares base, safety-aligned, and reasoning models, uncovering a reasoning-exclusive negative correlation we term \textit{Entropy-Gradient Inversion}.
Subsection 2.3\ref{sec:training_dynamics} tracks how this inversion emerges during SFT and strengthens under RL.

\subsection{Preliminaries: Mathematical Foundations of Gradient Influence and Entropy}

\label{subsec:preliminaries_entropy_gradient}

In this subsection, we formalize the core metrics used to quantify the internal gradient dynamics and external prediction uncertainty of LLMs, which lay the foundation for our analysis of the Entropy-Gradient Inversion phenomenon.

\textbf{Quantifying LLM prediction via output entropy and logit gradient.} To quantify internal response of the model and its relationship with prediction uncertainty, we compute the gradient dynamics of the primary attention projection layers and the corresponding output entropy. For token $t_i$ at layer $l$, we derive the gradient matrix $G_{X,l}$ where $X \in \{Q, K, V, O\}$ projection heads. The intensity of these updates is measured using the nuclear norm, defined as the $l_1$ norm of the matrix's singular values $\sigma_{j}$, and the overall influence $\bar{I}_{t_i}$ is obtained by averaging these norms across all $L$ layers:
\begin{equation}
s_{X, l}=\left\| G_{X, l}\right\| _{*}=\sum_{j}\left|\sigma_{j}\right|, \quad \bar{I}_{t_i} = \frac{1}{L} \sum_{l=1}^{L} \sum_{X \in \{Q,K,V,O\}} s_{X,l}.
\end{equation}
Simultaneously, we measure the prediction uncertainty using the Shannon entropy $E_i$ of the predictive distribution $P(v | x_{1:i-1})$ over the vocabulary $V$:
\begin{equation}
E_i = -\sum_{v \in V} P(v | x_{1:i-1}) \cdot \log_2 P(v | x_{1:i-1}).
\end{equation}
\textbf{Estimating the entropy and gradient through correlation.} The structural relationship between internal gradient intensity and external uncertainty is formally characterized using the Spearman $\rho = \frac{\sum_{i=1}^{N} (R(E_i) - \bar{R}(E))(R(I_i) - \bar{R}(I))}{\sqrt{\sum_{i=1}^{N} (R(E_i) - \bar{R}(E))^2 \sum_{i=1}^{N} (R(I_i) - \bar{R}(I))^2}}$, where $\bar{R}(E)$ and $\bar{R}(I)$ are the mean ranks of Entropy $E_i$ and Gradient Influence $\bar{I}_{t_i}$, and Pearson $r = \frac{\sum_{t} (\bar{I}_t - \bar{I})(E_i - \bar{E})}{\sqrt{\sum_{t} (\bar{I}_t - \bar{I})^2 \sum_{t} (E_i - \bar{E})^2}}$ correlation coefficient \citep{spearman1961proofspearman,pearson1896vii} between the logit overall inference $\bar{I}_t$ and entropy $E_i$ for a set of $n$ unique tokens. This correlation serves as an indicator that reflects the relationship between model's internal gradient representations and its external outputs entropy as the model's reasoning capability.

\subsection{Entropy-Gradient Inversion: The Reasoning Fingerprint in LRMs}
\label{subsec:entropy_gradient_inversion_subsec}

With quantitative metrics defined in Section \ref{subsec:preliminaries_entropy_gradient}2.1, we evaluate token-level prediction entropy and logit gradient magnitude correlations across base, safety, and reasoning models. 

\textbf{Preliminary setup.} To validate the uniqueness of the Entropy-Gradient Inversion phenomenon in LRMs, we conduct controlled empirical analysis variants with consistent model architecture but different training objectives: Qwen2.5-7B \citep{qwen2.5} specifically serves as the \textbf{Base Model}, Low-Rank Adaptation \citep{hu2021loralowrankadaptationlarge} fine-tunes the model on Safety-Tuned Dataset \citep{bianchi2024safetytuned} which yields the \textbf{Safe Model}, and DeepSeek-R1-Distill-Qwen-7B \citep{guo2025deepseek} represents the \textbf{Reasoning Model}. 
To evaluate correlation across diverse task distributions, we use three datasets: ARC-C \citep{clark2018thinksolvedquestionansweringARC} as \textbf{Base Samples}, hh-rlhf \citep{Bai2022TrainingAHhh-rlhf} as \textbf{Safety Samples}, and OpenThoughts-114k-math \citep{guha2025openthoughtsdatarecipesreasoning} as \textbf{Reasoning Samples}. Methods hyperparameter settings of preliminary experiments are shown in Appendix B\ref{app:baselines_hyper_setting}.


\input{multi_plots/correlation_samples_qwen}



\textbf{Entropy-Gradient Inversions serves as a fingerprint that represents reasoning capabilities in LRMs.} As shown in Figure \ref{fig:correlation_samples_qwen} (Left), Reasoning Model exhibits a robust negative Spearman correlation of $\rho = -0.649$ on reasoning samples. In contrast, the Base model shows a weak negative correlation of $\rho = -0.171$, while the Safe Model demonstrates a positive correlation of $\rho = 0.148$. 
These results indicate that in non-reasoning models, high entropy necessitates a larger gradient update to rectify the prediction, whereas reasoning models have internalized the structural logic of reasoning paths, allowing high-entropy tokens to maintain a low gradient impact on the model's weights. 
This negative correlation indicates that LRMs have internalized structural reasoning logic, enabling high-entropy ``slow thinking'' tokens to maintain minimal gradient impact on model weights.

\textbf{\textit{Key Finding 1.}} \textit{\textbf{The {Entropy-Gradient Inversion} phenomenon exists as a unique reasoning ``fingerprint'' in LRMs}. Only reasoning models, featuring emergent long chain-of-thought and self-reflection capabilities, exhibit a robust negative correlation between token entropy and gradient magnitude, while base and safety-aligned models show weak negative or even positive correlations. 
}

\subsection{Training Dynamics: The Evolution of Entropy-Gradient Inversion via SFT and RL}

\label{sec:training_dynamics}
\input{multi_plots/entropy_gradient_inversion_step}

Having established Entropy-Gradient Inversion as a unique fingerprint of mature LRMs, a natural question arises: \textit{at which stage of reasoning-enhancement training does this fingerprint emerge, and how is it progressively shaped?}
This subsection tracks dynamic entropy-gradient correlation changes across standard LRM pipeline using SFT and RL.



\textbf{Training Dynamics experiments setup.} The emergence of the inversion phenomenon is tracked through the R1 training pipeline, involving LoRA as SFT method \citep{ouyang2022traininglanguagemodelsfollow} and GRPO \citep{guo2024deepseekmath} with OpenThoughts-114k-math \citep{openr1} and OpenR1-Math-220k \citep{openr1} as the training dataset. During SFT, the model minimizes the negative log-likelihood of reasoning trajectories $\mathcal{D}$ using $\mathcal{L}_{\text{SFT}}(\theta) = -\mathbb{E}_{(x,y) \sim \mathcal{D}} \left[ \sum_{t=1}^{|y|} \log \pi_\theta(y_t \mid x, y_{<t}) \right]$.
Following SFT, GRPO refines the policy by maximizing a clipped surrogate objective using group-based advantage estimation. The core optimization objective of GRPO is defined as:

\begin{equation}
\begin{aligned}
    &\mathcal{J}_{\mathrm{GRPO}}(\theta) = \mathbb{E} \Biggl[  \frac{1}{m}\sum_{i=1}^m \\
    & \min\Bigl(r_{i,t} A_{i},\, \mathrm{clip}\bigl(r_{i,t}; 1-\epsilon, 1+\epsilon\bigr) A_{i}\Bigr) - \beta\, D_{\mathrm{KL}}\left(\pi_\theta\Vert\pi_{\mathrm{ref}}\right) \Biggr],
\end{aligned}
\label{equ:grpo_func}
\end{equation}
where $r_{i,t}$ is the probability ratio between the current policy and the reference policy, group advantage $A_i$ can be derived by reward $R_i$ minus the group mean $A_i = \frac{R_i - \frac{1}{G}\sum_{j=1}^G R_j}{\sqrt{\frac{1}{G}\sum_{j=1}^G \left(R_j - \frac{1}{G}\sum_{k=1}^G R_k\right)^2}}$. $r_{i,t}$ is the probability ratio between current and reference policy, $G$ is the group size, $\epsilon$ is the clipping threshold, and $\beta D_{\mathrm{KL}}$ is the KL divergence penalty to prevent policy drift from the reference model.

\textbf{Entropy-Gradient Inversion Evolution through SFT and RL stages.} The training dynamics of models are visualized in Figure \ref{fig:entropy_gradient_inversion_step_qwen} (Left). During SFT, the correlation shifts from the base level of $-0.171$ to approximately $-0.308$ within the first 200 steps, eventually reaching $-0.494$ at step 8000. In the subsequent RL stage using GRPO, the model further strengthens this inversion effect during its search for optimal reasoning paths, before converging to a final Spearman correlation of $-0.556$.

\textbf{Evolution and Convergence through RL stages.} Figure \ref{fig:entropy_gradient_inversion_step_qwen} (Right) displays the pure reinforcement learning method same as the R-Zero pipeline. The RL trajectory exhibits initial instability during the first 300 steps, with correlation oscillations driven by the reinforcement learning exploration process. We refer to this phase as the initial exploration of RL training, which eventually converges to a Spearman coefficient of $-0.318$. This progression suggests that RL reinforces the inversion to solidify the model's slow thinking mechanism. Detailed entropy-gradient results across different model architectures and formula derivation are provided in Appendix C and D.


\textit{\textbf{Key Finding 2.} \textbf{The Entropy-Gradient Inversion phenomenon emerges rapidly during SFT with a clear ``phase transition'' in the early training steps, and is further strengthened and solidified by subsequent RL training}. While pure RL without SFT warm-up suffers from training instability and fails to achieve strong inversion, the standard SFT+RL pipeline drives the correlation to a significantly deeper negative state, which directly aligns with the model's superior reasoning performance.}

%% file: multi_plots/correlation_samples_qwen.tex
\begin{figure*}[t]
    \centering
      \centering
    \begin{minipage}{1\textwidth}
      \centering
    \includegraphics[width=0.55\textwidth]{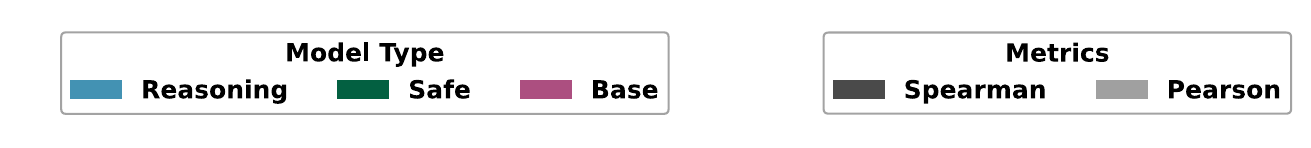}
    \label{fig:XAI_Samples_correlation_legend}
  \end{minipage}
  \vspace{-2em}
  \vskip\baselineskip 
    \subfigure{
        \includegraphics[width=0.31\textwidth]{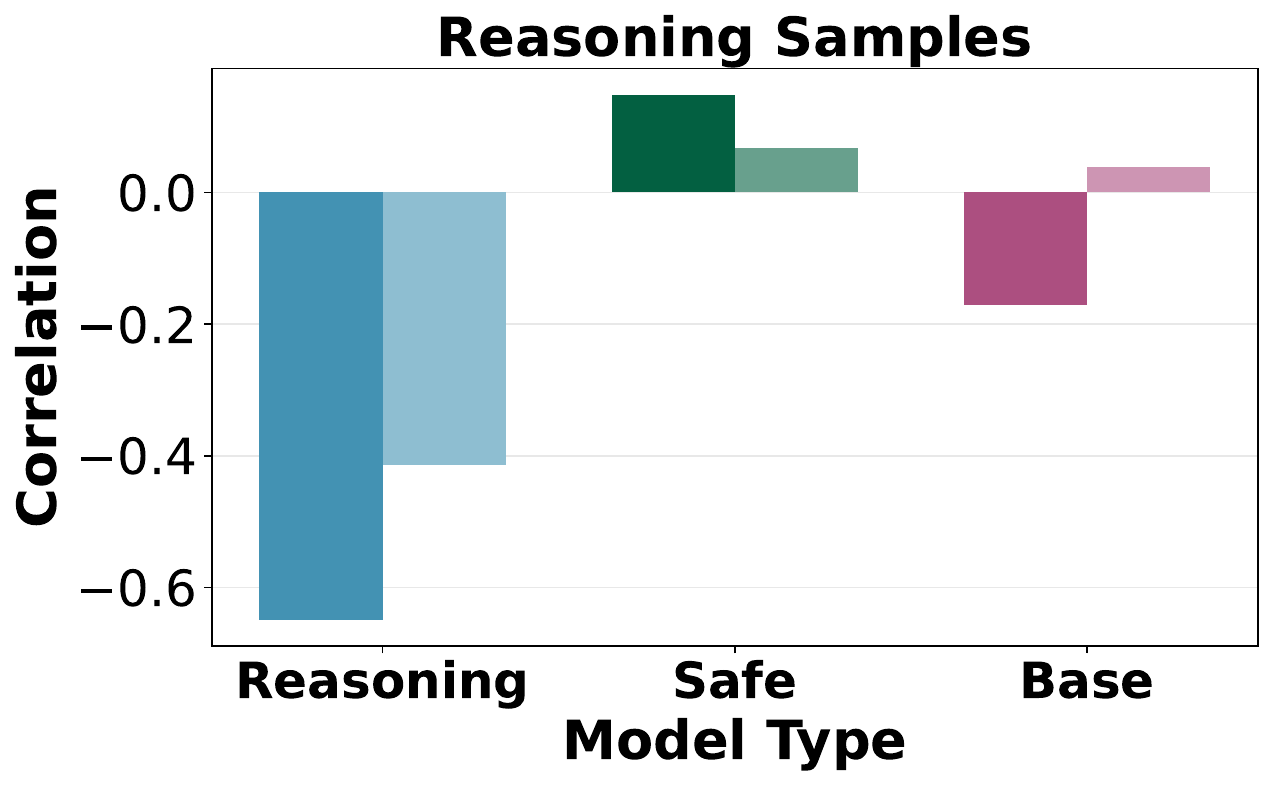}
        \label{fig:entropy_logit_gradient_reasoning_samples_Qwen_Combined}
    } 
    \hfill  
    \subfigure{
        \includegraphics[width=0.31\textwidth]{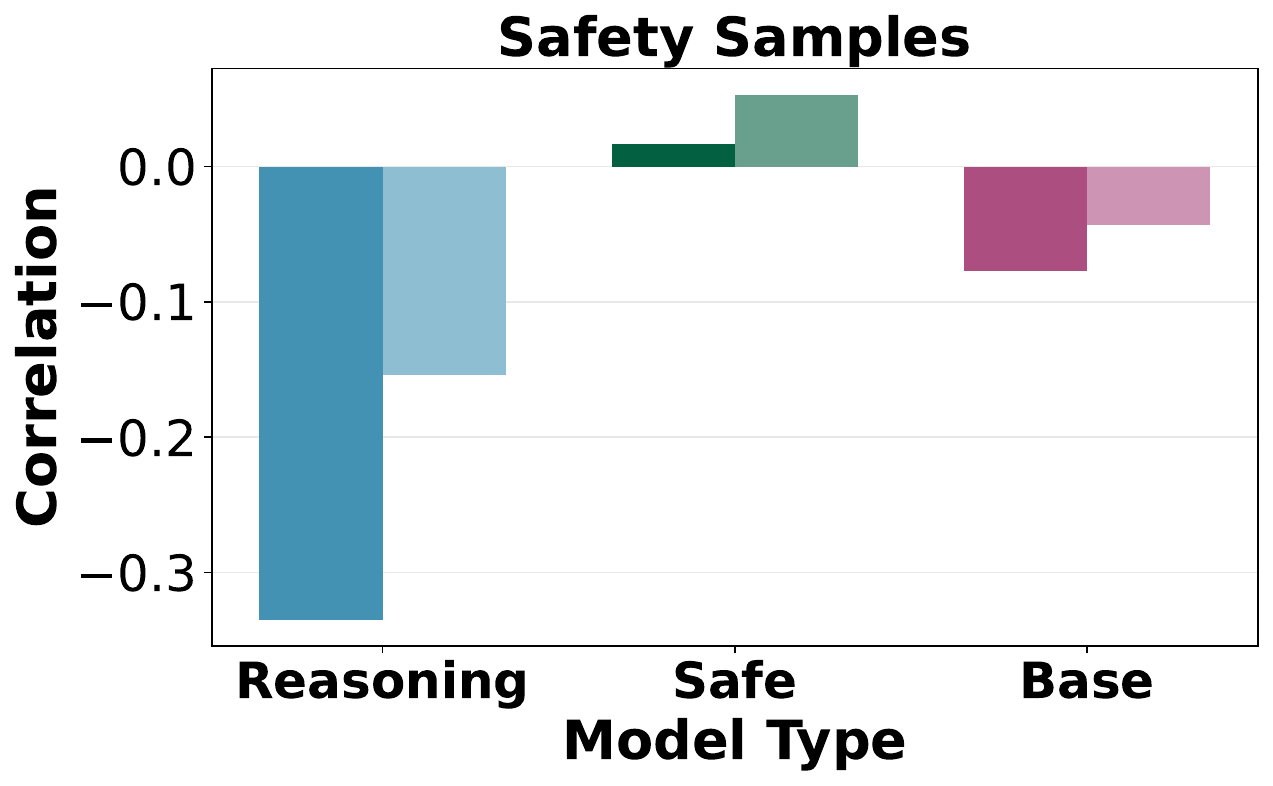}
        \label{fig:entropy_logit_gradient_safety_samples_Qwen_Combined}
    }
    \hfill  
    \subfigure{
        \includegraphics[width=0.31\textwidth]{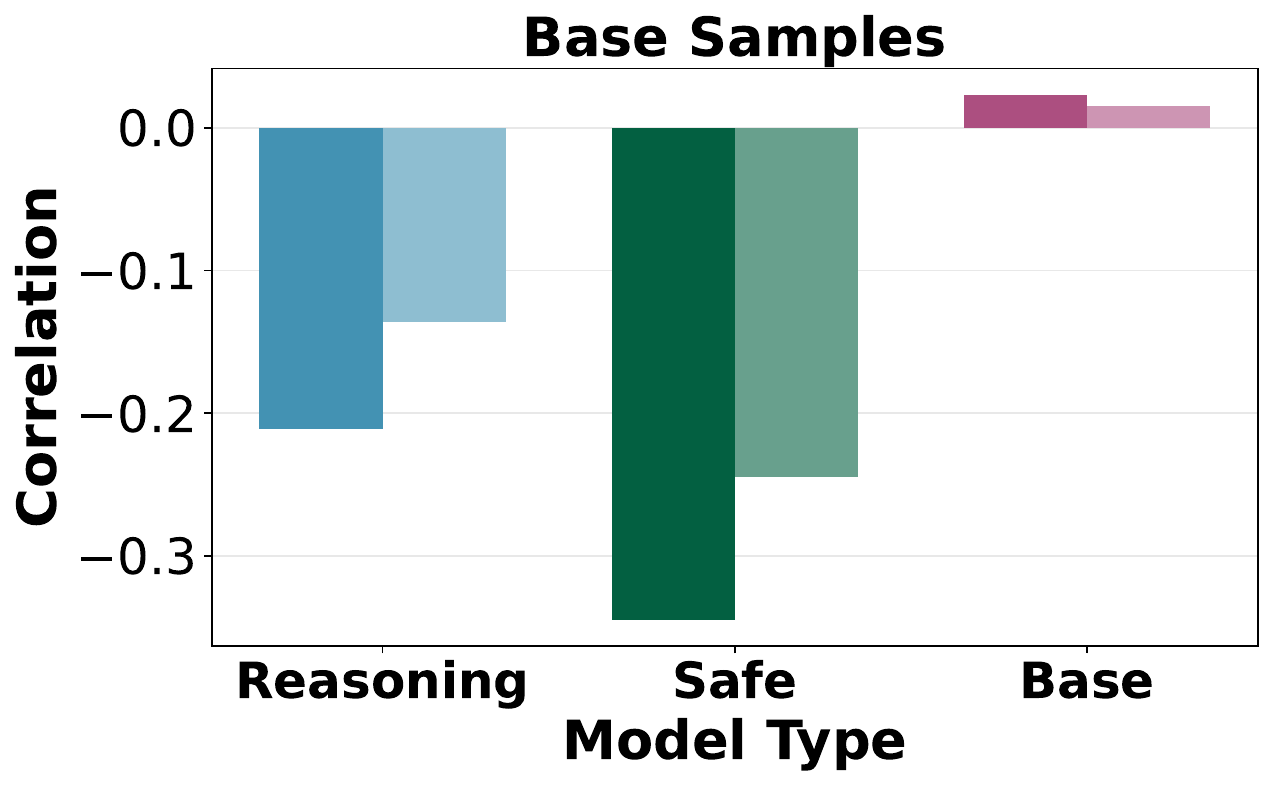}
        \label{fig:entropy_logit_gradient_base_samples_Qwen_Combined}
    }
    
    \vspace{-0.5em}
    \caption{\textbf{Spearman correlation between logit gradient nuclear norm and token entropy across different model types on Qwen2.5-7B family}. The subfigures illustrate the correlation analysis conducted on different data distributions, specifically showing \textbf{Left}: \textit{Reasoning Samples}, \textbf{Middle}: \textit{Safety Samples}, and \textbf{Right}: \textit{Base Samples}. In each case, we evaluate three distinct model variants (Reasoning, Safety, and Base models) to demonstrate the consistent monotonic relationship between gradient sensitivity and predictive uncertainty across both diverse tasks and model alignment stages.}
    \label{fig:correlation_samples_qwen}
\end{figure*}

%% file: multi_plots/entropy_gradient_inversion_step.tex
\begin{figure}[t]
    
      \centering
      \subfigure{
        \includegraphics[width=0.47\columnwidth]{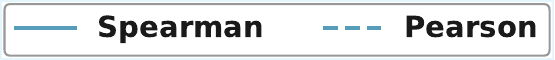}
        \label{fig:pipeline_legend}
        } 
  \vspace{-1.5em}
  \vskip\baselineskip 
    \subfigure{
        \includegraphics[width=0.47\columnwidth]{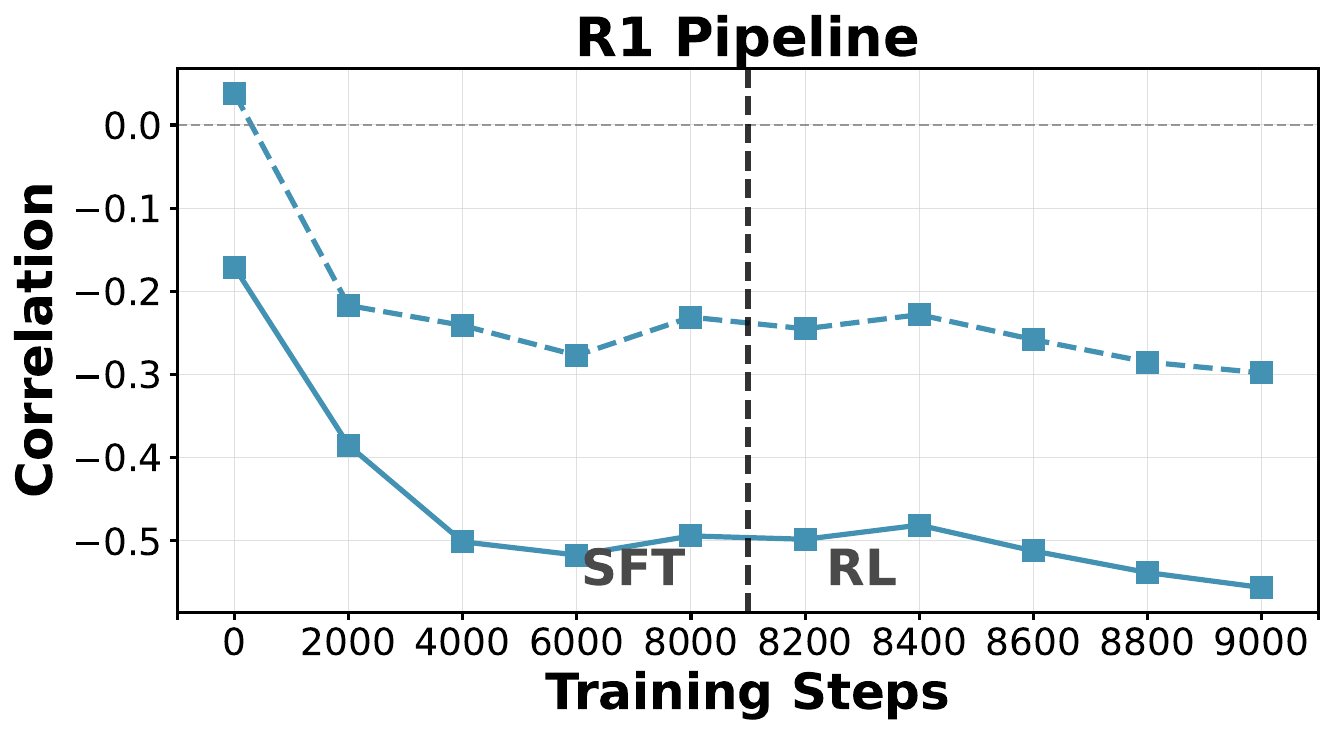}
        \label{fig:r1_pipeline_qwen25_7b}
    } 
    \subfigure{
        \includegraphics[width=0.47\columnwidth]{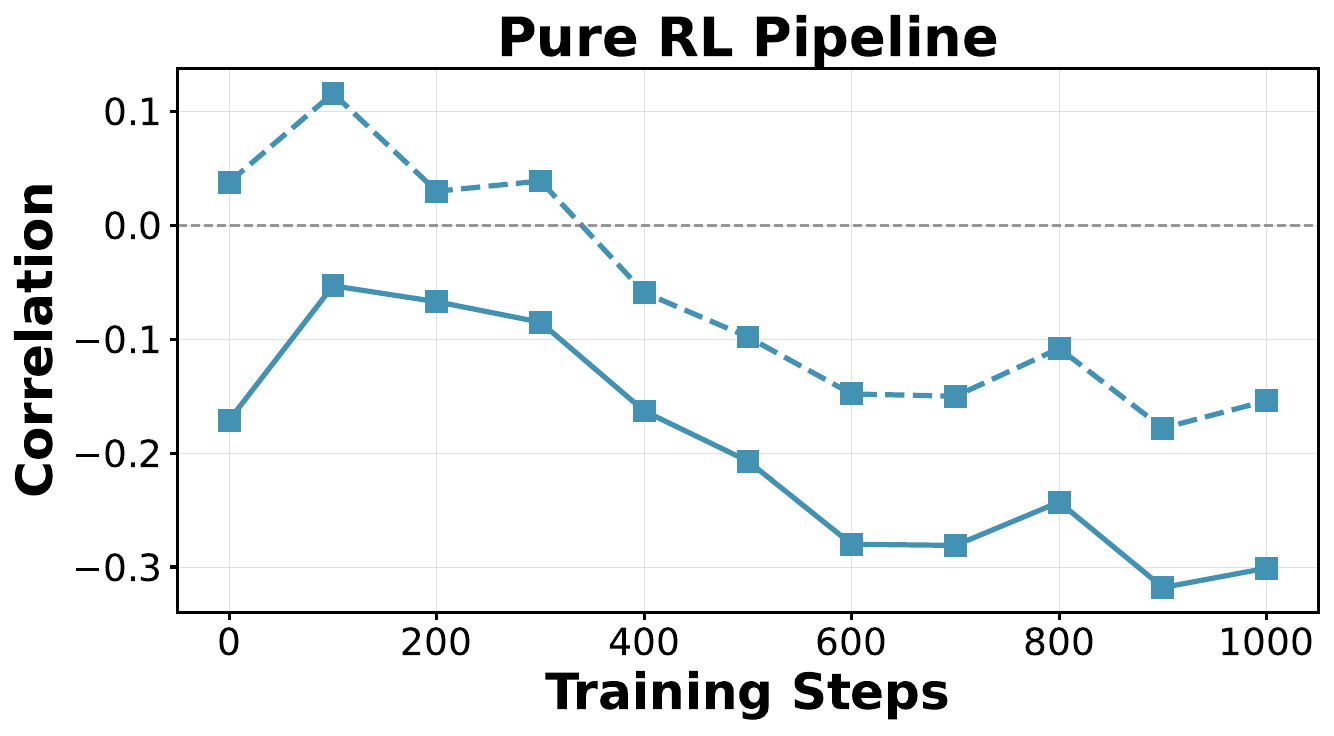}
        \label{fig:pure_rl_pipeline_qwen25_7b}
    }
    
    \vspace{-0.5em}
    \caption{\textbf{Comparative analysis of correlation variance across three training methodologies using Qwen2.5-7B}. \textbf{Left}: The standard DeepSeek-R1 pipeline, consisting of sequential SFT on reasoning data followed by GRPO-based reinforcement learning, every $2000$ steps in SFT stage has the same width compared with every $200$ steps in RL stage. \textbf{Right}: A pure RL pipeline where GRPO is applied directly to the base model without an SFT warm-up phase.}
    \label{fig:entropy_gradient_inversion_step_qwen}
\end{figure}

%% file: sections/method.tex
\section{CorR-PO: Leveraging Entropy-Gradient Inversion to Boost LRM Reasoning}

\label{sec:application_corrpo}

Given that the Entropy-Gradient Inversion tightly couples with strong reasoning capability yet only gradually realizes across post-training, a natural research question arises: rather than waiting for this geometric signature to emerge passively as a by-product of outcome-level optimization, \textit{can we explicitly embed it as a per-response regularization into the reinforcement learning reward, so as to steer LRMs toward reasoning-aligned latent structures from the outset of training and achieve more stable state-of-the-art reasoning performance?} 

To address this, we draw insights from our analysis of Entropy-Gradient Inversion.
We propose a simple yet effective technique to improve LRMs’ reasoning performance, and introduce CorR-PO, an Entropy-Gradient Correlation guided reinforcement learning method for LLMs that augments GRPO \citep{guo2024deepseekmath} with a specialized correlation regularization reward.
Rather than rewarding diversity in output space, CorR-PO introduces an intrinsic penalty that actively suppresses geometric configurations associated with fast thinking during reasoning steps. 

\subsection{Preliminaries}
\paragraph{Autoregressive LLMs.} Let $\mathcal{X}$ be a set of prompts and $\mathcal{Y}$ be the response. For $x\in\mathcal{X}$, a response is the token sequence $y=(y_1,\dots,y_L)\in\mathcal{Y}$. An autoregressive LLM with parameters $\theta$ is defined as:
\begin{equation}
\pi_\theta(y\mid x) \;=\; \prod_{t=1}^L \pi_\theta\left(y_t \mid x, y_{<t}\right).
\label{eq:llm}
\end{equation}
\paragraph{Reinforcement Learning in LLMs.} For each prompt $x$ we generate a set of $m$ candidate responses $\{y^{(i)}\}_{i=1}^m$ from a fixed behavior policy $\pi_{\theta^{\mathrm{old}}}$ using auto-regressive sampling.
Each response receives a \emph{base} scalar reward $R_i\in\mathbb{R}$, e.g., a verifiable pass/fail or a task-specific score.

\paragraph{Group-Relative Policy Optimization.} GRPO \citep{guo2024deepseekmath} dispenses with a learned critic and estimates groupwise advantages by standardizing rewards within the group. Specifically, it  computes the group mean reward as $\text{mean}(R_1, \dots, R_G) = \frac{1}{m}\sum_{i=1}^m R_i$, then calculates the group standard deviation $\text{std}(R_1, \dots, R_G) = \sqrt{\frac{1}{m}\sum_{i=1}^m (R_i - \text{mean}(R_1, \dots, R_G))^2 + \varepsilon}$ with $\varepsilon>0$ for numerical stability, and finally derives the advantage for each response as $A_i = \frac{R_i - \text{mean}(R_1, \dots, R_G)}{\text{std}(R_1, \dots, R_G)} \label{eq:grpo-adv}$. Let:
\begin{gather}
r_{i,t} \;=\; \frac{\pi_\theta\left(y^{(i)}\mid x\right)}{\pi_{\theta^{\mathrm{old}}}\left(y^{(i)}\mid x\right)}, \\
\mathrm{clip}\left(u;1-\epsilon,1+\epsilon\right)\;=\;\min\bigl\{\max\{u,1-\epsilon\},\,1+\epsilon\bigr\},
\label{eq:ratio-clip}
\end{gather}
and let $D_{\mathrm{KL}}\left(\pi_\theta \Vert\pi_{\mathrm{ref}}\right)$ be a per-prompt KL penalty to a reference policy of Base model. GRPO objective is derived into Equation \ref{equ:grpo_func}. 
CorR-PO modifies only rewards construction, with PPO-style clipping and KL control unchanged.

\subsection{CorR-PO: Regularize Inversion Correlation via Group Policy Optimization}

Utilizing the correlation feature of Entropy-Gradient Inversion in Section 2, \textbf{Correlation-Regularized Group Policy Optimization} (\textbf{CorR-PO}) transcends the traditional reliance on singular outcome rewards by directly embedding our discovered geometric heuristic rule into the GRPO framework. To operationalize the geometric properties of slow thinking, we systematically evaluate the correlation dynamics between predictive uncertainty and internal gradient across the reasoning trajectory.

\textbf{Uncertainty Prediction via  Step Average Entropy.} The generated sequence of tokens is segmented into $N$ distinct reasoning sentences, where $s_i$ denotes the collection of tokens belonging to the $i$-th sentence. For each sentence $i \in \{1, 2, \dots, N\}$, we compute the Step Average Entropy ($E_i$). Following the phenomenon uncovered in Section 2, Step Average Entropy quantifies the model's predictive uncertainty during the reasoning phase and is calculated as the average token-level predictive entropy across all tokens within the sentence:
\begin{gather}
E_i = \frac{1}{|s_i|} \sum_{t \in s_i} H_i = -\sum_{v \in V} P(v | x_{1:i-1}) \cdot \log_2 P(v | x_{1:i-1}).
\end{gather}

\textbf{Internal Gradient Influence.} To evaluate the structural properties of the model's latent space, we derive the overall Gradient Influence ($\bar{I}_t$) by computing nuclear norm of the gradient matrices $G_{X,l}^{(t)}$ for each token $t$ across all $L$ layers:
\begin{gather}
    s_{X, l}^{(t)} = \left\| G_{X, l}^{(t)}\right\| _{*} = \sum_{j}\left|\sigma_{j}\right|, \
    \bar{I}_t = \frac{1}{L} \sum_{l=1}^{L} \sum_{X \in \{Q,K,V,O\}} s_{X,l}^{(t)}. 
\end{gather}

\textbf{Correlation Regularization Reward.} After processing the full reasoning sequence, we construct two length-$N$ arrays to characterize reasoning dynamics: a step entropy array $E = [E_1, E_2, \dots, E_N]$ for quantifying predictive uncertainty per reasoning step and a gradient influence array $I = [\bar{I}_1, \bar{I}_2, \dots, \bar{I}_N]$ for measuring internal parameter update intensity. We use the Spearman rank correlation \citep{spearman1961proofspearman} $\rho_{E,I}$ to capture their relationship, which relies on ranks rather than raw values, making it robust to outliers and effective at detecting monotonic associations, which aligns with our goal of identifying the inversion signature.

To compute $\rho_{E,I}$, we first convert raw values in $E$ and $I$ to their corresponding ranks, denoted $R(E_i)$ and $R(I_i)$ for step $i$, where $\bar{R}(E)$ and $\bar{R}(I)$ represent the mean ranks of  $E_i$ and  $\bar{I}_{t_i}$, respectively. The Spearman correlation, which can be formulated as the Pearson correlation of ranks, and the regularization reward can be derived as:
\begin{gather}
    \rho_{E,I} = \frac{\sum_{i=1}^{N} (R(E_i) - \bar{R}(E))(R(I_i) - \bar{R}(I))}{\sqrt{\sum_{i=1}^{N} (R(E_i) - \bar{R}(E))^2 \sum_{i=1}^{N} (R(I_i) - \bar{R}(I))^2}}, \\
    R_{corr} = -(1 + \rho_{E,I}).
    \label{equ:reward_corr}
\end{gather}


The offset $+1$ in the correlation-based regularization shifts $\rho_{E,I}\in[-1,1]$ into $R_{corr}\in[-2,0]$, yielding a one-sided non-positive penalty that never competes with the accuracy reward. By penalizing non-negative correlations $\rho_{E,I} \geq 0$, we steer the model's latent space toward structured ``slow thinking'' trajectories, reducing RL's reliance on costly external verifiers and stabilizing training.

\textbf{CorR-PO Objective.} The intrinsic correlation regulation $R_{corr}$ derived in Equation \ref{equ:reward_corr} is scaled by a hyperparameter $\lambda_{corr}$ and integrated with the rule-based ground-truth accuracy reward $R_{acc}$. The final sequence-level total reward is formulated as:
\begin{gather}
  R_{total} = R_{acc} + \lambda_{corr}R_{corr}=R_{acc} - \lambda_{corr} (1 + \rho_{E,I})  .
\end{gather}

By utilizing this correlation-regularized reward $R_{total}$ to compute the group-relative advantages $A_i$ (standardized within a group of size $G$), we optimize the policy using the modified objective function:

\begin{equation}
\begin{aligned}
    &\mathcal{J}_{\text{CorR-PO}}(\theta) = \mathbb{E} \Biggl[  \frac{1}{G} \sum_{i=1}^{G} \frac{1}{|o_i|} \sum_{t=1}^{|o_i|} \Biggl( \\
    & \min \left( r_{i,t} A_i, \text{clip}(r_{i,t}, 1 - \epsilon, 1 + \epsilon) A_i \right) - \beta D_{\text{KL}} \Biggr) \Biggr] .
\end{aligned}
\end{equation}



Here, $|o_i|$ is the sequence length of the $i$-th output, $r_{i,t}$ denotes the probability ratio of the active policy to the reference policy, and $\beta D_{KL}$ serves as a Kullback-Leibler divergence \citep{Kullback1951OnIA_KL} penalty to prevent reward hacking. This guided optimization leverages internal mechanistic signals to {augment domain-specific verifiers, alleviating the model’s dependence on external supervision while mitigating RL’s inherent optimization instability and enabling more stable state-of-the-art reasoning performance.}

\label{subsec:corr_po}

%% file: sections/main_experiment.tex
\section{Experiments}

\subsection{Experiment Setup}

\input{tables/math_performance_experiments/performance_qwen25_7b_math}

\textbf{Baselines.} We compare CorR-PO against the unaligned \textbf{Base} model and state-of-the-art methods \textbf{GRPO} \citep{guo2024deepseekmath}, \textbf{DAPO} \citep{yu2025dapoopensourcellmreinforcementdapo}, \textbf{Dr.GRPO} \citep{liu2025understandingr1zeroliketrainingcritical}, and \textbf{GSPO} \citep{zheng2025groupsequencepolicyoptimizationgspo}. Detailed baseline explanations and
recommended hyperparameter settings are listed in A and B. 
\textbf{Training Set.} Following RL pipeline in Section 2.3, we utilize OpenR1-Math-220k \citep{openr1} as training dataset.
\textbf{Evaluation.} We assess reasoning capabilities using Pass@1, Pass@16, and Major@16 metrics on \textbf{AIME24} \citep{aime_1983_2024}, \textbf{MATH500} \citep{lightman2023letmath500}, and \textbf{GSM8k} \citep{Cobbe21gsm8k}. \textbf{Models.} We adopt \textbf{Qwen2.5-7B-Math} \citep{yang2024qwen25mathtechnicalreportmathematical}, \textbf{Qwen2.5-14B} \citep{qwen2.5}, \textbf{Qwen3-4B} and \textbf{Qwen3-1.7B} \citep{yang2025qwen3technicalreport} for our primary experiments.

\subsection{CorR-PO Improves Reasoning Performance through Correlation Regulation}
\label{sec:main_exp_math_qwen25_7b}

\label{sec:main_exp_math_qwen25_14b}
\input{tables/math_performance_experiments/performance_qwen25_14b}
\textbf{CorR-PO achieves better reasoning performance compared with baseline methods.} Table \ref{tab:performance_qwen25_7b_math} presents the Pass@1, Pass@16, and Major@16 evaluation results on AIME24, MATH500, and GSM8k benchmarks, which cover mathematical reasoning and complex problem-solving scenarios, using Qwen2.5-7B-Math as the base model. CorR-PO achieves the highest average performance of $69.4$, outperforming the best baseline GSPO by $0.8$ percentage points and vanilla GRPO by $2.4$ percentage points, and consistently leads on core sub-metrics across all benchmarks, demonstrating its  ability to optimize model reasoning capability via the proposed correlation regularization.

\textbf{CorR-PO performs stably across model families and scales.} Table \ref{tab:performance_qwen25_14b} shows CorR-PO reaches an average performance of $72.9$ with Qwen2.5-14B as the base model, surpassing the top baseline Dr.GRPO by $1.4$ percentage points, achieving leading results on key sub-metrics and consistently maintaining superiority over all competing baselines. 
For additional experiments and evaluations on model sizes such as Qwen3-1.7B and Qwen3-4B \citep{yang2025qwen3technicalreport}, please refer to Appendix E.

\subsection{Training Dynamics}

\input{multi_plots/step_rl_corr_accuracy}

\textbf{Model Achieves Stronger Entropy-Gradient Inversion through CorR-PO.} As shown in Figure \ref{fig:step_rl_corr_accuracy} (Left), it illustrates the dynamic evolution of the Spearman correlation coefficient between token entropy and logit gradient nuclear norm across $1000$ training steps and we save the checkpoint every 100 steps for model evaluation during the training process. 
Compared with the state-of-the-art GRPO baseline, CorR-PO consistently drives the correlation from $-0.171$ to a more negative state of $-0.363$ at step $1000$, surpassing GRPO's $-0.301$, rapidly forming and steadily maintaining Entropy-Gradient Inversion features of reasoning models.

\textbf{Stronger Entropy-Gradient Inversion tends to represent better model reasoning performance.} As shown in Figure \ref{fig:step_rl_corr_accuracy} (Right), CorR-PO average performance increases to $70.1$ at step $1000$, outperforming the $66.0$ GRPO baseline score. This trend is positively correlated with the stronger Entropy-Gradient Inversion effect observed in the left subfigure. Full training dynamics experiment results across all benchmarks can be found in Appendix F.

\subsection{Hyperparameter Analysis}

\textbf{CorR-PO is robust to hyperparameter choices and benefits from sufficient correlation regularization.} As shown in Table \ref{tab:hyperparameter}, we conduct a comprehensive hyperparameter analysis of CorR-PO on the Qwen2.5-7B-Math base model, evaluating two learning rates of $1.0 \times 10^{-6}$ and $3.0 \times 10^{-6}$ and four correlation regularization coefficients $\lambda_{corr}\in\{0.05, 0.15, 0.25, 0.35\}$ across AIME24, MATH500, and GSM8k benchmarks. All eight configurations fall within a narrow average range from  $66.2$ to $69.4$, confirming that CorR-PO is not sensitive to specific hyperparameter choices. The $1.0 \times 10^{-6}$ learning rate yields tightly clustered results ranging from $66.2$ to $66.6$, whereas $3.0 \times 10^{-6}$ exposes a wider landscape where a sufficiently strong penalty is required: the optimal configuration with learning rate of $3.0 \times 10^{-6}$ with $\lambda_{corr}=0.35$ reaches $69.4\%$ average performance and tops $6$ of $9$ sub-metrics, confirming that a larger $\lambda_{corr}$ more aggressively enforces the Entropy-Gradient Inversion prior and translates the resulting geometric structure into reasoning gains.

\input{tables/hyperparameter/hyperparameter_analysis}

%% file: tables/math_performance_experiments/performance_qwen25_7b_math.tex
\begin{table*}[t]
\centering
\caption{Main results of CorR-PO and baseline methods on AIME24, MATH500, GSM8k with Qwen2.5-7B-Math as the base model. All numbers are percentage performances and the best performance among all methods on each dataset is highlighted in \textbf{bold}, while the second-best is marked with \underline{underline}. Average $\uparrow$ column indicate average performance across all benchmarks.}
\label{tab:performance_qwen25_7b_math}
\renewcommand{\arraystretch}{0.95} 
\resizebox{\textwidth}{!}{\begin{tabular}{l|ccc|ccc|ccc|c}
\toprule
\textbf{Datasets} & \multicolumn{3}{c|}{\textbf{AIME24}} & \multicolumn{3}{c|}{\textbf{MATH500}} & \multicolumn{3}{c|}{\textbf{GSM8k}} & \\
 \midrule
\textbf{Metrics} & Pass@1 & Pass@16 & Major@16 & Pass@1 & Pass@16 & Major@16 & Pass@1 & Pass@16 & Major@16 & Average$\uparrow$ \\
\midrule
\textbf{Base} & 10.0 & 40.0 & 20.0 & 60.4 & 90.8 & 79.8 & 82.3 & 97.3 & 90.2 & 63.4 \\
\textbf{GRPO} & 16.7 & \underline{50.0} & 23.3 & \underline{71.4} & 90.6 & 80.2 & 82.8 & \underline{97.5} & 90.6 & 67.0 \\
\textbf{DAPO} & \textbf{26.7} & 40.0 & \underline{26.7} & 71.0 & 91.0 & 80.2 & \underline{84.2} & 97.3 & \underline{91.9} & 67.7 \\
\textbf{Dr.GRPO} & 16.7 & \underline{50.0} & \underline{26.7} & 69.6 & 90.0 & 79.2 & 83.7 & 96.9 & 89.6 & 66.9 \\
\textbf{GSPO} & \underline{23.3} & \underline{50.0} & \underline{26.7} & 70.8 & \underline{91.4} & \underline{81.6} & \textbf{85.5} & 97.1 & 91.1 & \underline{68.6} \\
\rowcolor{lightgray!25} \textbf{CorR\textendash PO} & \underline{23.3} & \textbf{56.7} & \textbf{26.7} & \textbf{72.6} & \textbf{91.4} & \textbf{80.6} & 83.7 & \textbf{97.6} & \textbf{91.9} & \textbf{69.4} \\
\bottomrule
\end{tabular}}
\end{table*}

%% file: tables/math_performance_experiments/performance_qwen25_14b.tex
\begin{table*}[t]
\centering
\caption{Main results of CorR-PO and baseline methods on AIME24, MATH500, GSM8k with Qwen2.5-14B as the base model. All numbers are percentage performances and the best performance among all methods on each dataset is highlighted in \textbf{bold}, while the second-best is marked with \underline{underline}. Average $\uparrow$ column indicate average performance across all benchmarks.}
\label{tab:performance_qwen25_14b}
\renewcommand{\arraystretch}{0.9} 
\resizebox{\textwidth}{!}{\begin{tabular}{l|ccc|ccc|ccc|c}
\toprule
\textbf{Datasets} & \multicolumn{3}{c|}{\textbf{AIME24}} & \multicolumn{3}{c|}{\textbf{MATH500}} & \multicolumn{3}{c|}{\textbf{GSM8k}} & \\
 \midrule
\textbf{Metrics} & Pass@1 & Pass@16 & Major@16 & Pass@1 & Pass@16 & Major@16 & Pass@1 & Pass@16 & Major@16 & Average$\uparrow$ \\
\midrule
\textbf{Base} & 10.0 & 43.3 & 16.7 & 74.2 & {85.4} & {79.8} & 90.8 & 98.2 & 94.0 & 65.8 \\
\textbf{GRPO} & 16.7 & 50.0 & 43.3 & 73.8 & 90.0 & 78.2 & 91.7 & \underline{98.3} & 93.9 & 70.7 \\
\textbf{DAPO} & \textbf{23.3} & 50.0 & 40.0 & 71.0 & 89.8 & 76.0 & \underline{93.9} & 97.9 & \textbf{95.5} & 70.8 \\
\textbf{Dr.GRPO} & 13.3 & \textbf{56.7} & \underline{46.7} & 74.8 & 90.2 & \underline{80.3} & 91.4 & 97.5 & 94.1 & \underline{71.5} \\
\textbf{GSPO} & 16.7 & 50.0 & 43.3 & \underline{75.0} & \underline{90.4} & 76.8 & 92.6 & 97.8 & \underline{94.2} & 70.8 \\
\rowcolor{lightgray!25} \textbf{CorR\textendash PO} & \underline{20.0} & \underline{53.3} & \textbf{50.0} & \textbf{75.6 }& \textbf{90.6} & \textbf{81.2} & \textbf{93.6} & \textbf{98.3} & 93.9 & \textbf{72.9} \\
\bottomrule
\end{tabular}}
\end{table*}

%% file: multi_plots/step_rl_corr_accuracy.tex
\begin{figure}[t]
    \centering

    \subfigure{
        \includegraphics[width=0.47\columnwidth]{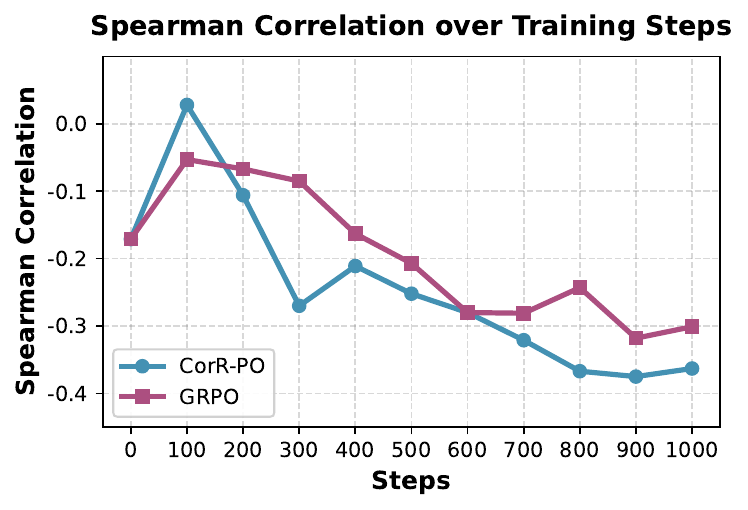}
        \label{fig:spearman_correlation_over_steps}
    }
    \subfigure{
        \includegraphics[width=0.47\columnwidth]{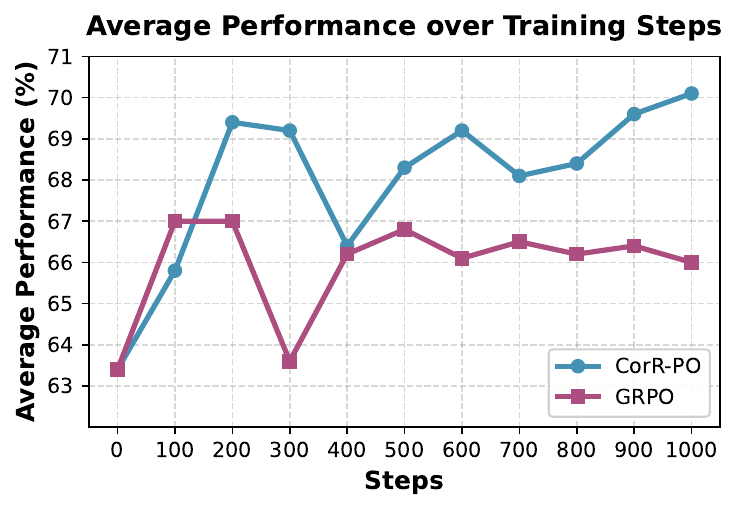}
        \label{fig:accuracy_over_steps}
    } 

    \vspace{-0.5em}
    \caption{\textbf{Spearman correlation (Left) and average performance (Right) over training steps. }Stronger entropy-gradient inversion is positively correlated with model reasoning performance. As shown in the right subfigure, CorR-PO achieves better performance across multiple reasoning benchmarks compared with GRPO as baseline method.}
    \label{fig:step_rl_corr_accuracy}
\end{figure}

    


%% file: tables/hyperparameter/hyperparameter_analysis.tex
\begin{table*}[t]
\centering
\caption{Hyperparameter analysis of CorR-PO with varying learning rates and correlation regularization coefficient $\lambda_{corr}$ on AIME24, MATH500, and GSM8k with Qwen2.5-7B-Math as the base model. All numbers are percentage performances, and the Average $\uparrow$ column indicates the average performance across all benchmarks. The best performance on each column is highlighted in \textbf{bold}, while the second-best is marked with \underline{underline}.}
\label{tab:hyperparameter}
\renewcommand{\arraystretch}{0.8} 
\resizebox{\textwidth}{!}{\begin{tabular}{l|ccc|ccc|ccc|c}
\toprule
\textbf{Datasets} & \multicolumn{3}{c|}{\textbf{AIME24}} & \multicolumn{3}{c|}{\textbf{MATH500}} & \multicolumn{3}{c|}{\textbf{GSM8k}} & \\
 \midrule
\textbf{Metrics} & Pass@1 & Pass@16 & Major@16 & Pass@1 & Pass@16 & Major@16 & Pass@1 & Pass@16 & Major@16 & Average$\uparrow$ \\
\midrule
 & \multicolumn{9}{c}{\textbf{learning rate = $1.0 \times 10^{-6}$}} & \\
\midrule
$\lambda$=0.05 & 13.3 & \underline{50.0} & \underline{23.3} & 70.4 & 90.2 & 79.4 & \underline{83.7} & 97.3 & \underline{90.0} & 66.4 \\
$\lambda$=0.15 & 16.7 & 43.3 & \textbf{26.7} & 69.8 & \underline{91.2} & 80.0 & 82.8 & 96.9 & 89.6 & 66.3 \\
$\lambda$=0.25 & 16.7 & \underline{50.0} & 20.0 & 71.0 & 90.8 & \underline{81.0} & 83.2 & 97.0 & 89.8 & 66.6 \\
$\lambda$=0.35 & \textbf{26.7} & 43.3 & 16.7 & 69.8 & 90.0 & 79.2 & 83.2 & 96.9 & 89.6 & 66.2 \\
\midrule
 & \multicolumn{9}{c}{\textbf{learning rate = $3.0 \times 10^{-6}$}} & \\
\midrule
$\lambda$=0.05 & 13.3 & 40.0 & \underline{23.3} & 69.8 & 90.4 & \textbf{81.2} & \textbf{83.8} & \underline{97.4} & 89.8 & 65.4 \\
$\lambda$=0.15 & 20.0 & 46.7 & \textbf{26.7} & \underline{71.4} & 90.2 & 80.2 & 82.4 & 97.0 & 89.8 & \underline{67.2} \\
$\lambda$=0.25 & 20.0 & 46.7 & \underline{23.3} & 71.0 & 90.8 & 79.6 & 83.1 & 97.2 & 88.9 & 66.7 \\
\rowcolor{lightgray!25} $\lambda$=0.35 & \underline{23.3} & \textbf{56.7} & \textbf{26.7} & \textbf{72.6} & \textbf{91.4} & 80.6 & \underline{83.7} & \textbf{97.6} & \textbf{91.9} & \textbf{69.4} \\
\bottomrule
\end{tabular}}
\end{table*}

%% file: sections/related_work.tex
\section{Related Work}
\label{sec:related_work}

\paragraph{Reasoning Model Interpretability.}
Early LLM reasoning interpretability research established frameworks via mechanistic circuit analysis \citep{olah2020zoom}, attention visualization \citep{abnar2023transformer}, and causal neural computation abstraction \citep{geiger2023causal}, decoding inference’s structural basis \citep{nanda2023mechanistic,saphra2024mechanistic}. Subsequent works explored reasoning dynamics from multi-dimensional views, identifying core phenomena: thinking tokens’ mutual information peaks \citep{qian2025demystifyingreasoningdynamicsmutual}, dual-process layer-wise gradient discrepancies \citep{wei2022chain,han2023multi,zhao2024survey,li-etal-2025-happened}, and syllogistic inference causal circuits \citep{li2025reasoning}. Extensive studies quantified reasoning uncertainty \citep{malinin2020predictive}, and linked reasoning ability to parameter gradient sensitivity \citep{wang2023towards,yang2026reasonanyincorporatingreasoningcapability,Yang_Wang_Zhuang_Chen_Zeng_2026}, with cognitive theories \citep{kahneman2011thinking} and benchmarks \citep{clark2018thinksolvedquestionansweringARC,hendrycks2021measuring} as evaluation foundations.

\paragraph{Reinforcement Learning with Verifiable Rewards.}
LLM alignment via reinforcement learning originated from the RLHF paradigm \citep{schulman2017proximalpolicyoptimizationalgorithmsppo,ouyang2022training,Bai2022TrainingAHhh-rlhf,bai2022constitutional}, establishing reward modeling and policy optimization core pipeline; variants like DPO \citep{rafailov2024directpreferenceoptimizationlanguagedpo} and PPO \citep{schulman2017proximalpolicyoptimizationalgorithmsppo} simplified training and improved stability. As a task-specific branch, RLVR targets formally verifiable reasoning tasks \citep{hendrycks2021measuring,Cobbe21gsm8k,aime_1983_2024}, achieving math and commonsense reasoning breakthroughs \citep{lightman2023letmath500,yang2024qwen25mathtechnicalreportmathematical} via iterative RL on standardized benchmarks \citep{guo2024deepseekmath,jaech2024openai}. Recent works further optimized RLVR efficiency, identified high-entropy minority tokens as optimization drivers \citep{wangbeyond20258020rulehighentropyminority}, proposed gradient-guided exploration \citep{liang2025llmsguideexplorationgradientguidedGRL} and novelty-aware learning to prevent entropy collapse \citep{zhou2026evolvinglanguagemodelslabels}.

%% file: sections/conclusions.tex
\section{Conclusion and Discussion}


This work addresses two challenges in large reasoning model (LRM) research: the gap between token-level behavioral analysis and internal reasoning mechanisms, and the instability of reinforcement learning (RL) for reasoning optimization relying on external verifiers. We define Entropy-Gradient Inversion, a negative correlation between token entropy and logit gradients, as a geometric fingerprint for LRM reasoning capability, and characterize its evolution across SFT and RL stages. Building on this, we propose CorR-PO, which embeds this signature into RL reward regularization. Extensive experiments across benchmarks and model scales show CorR-PO outperforms baselines, delivering novel mechanistic insights into slow thinking and a practical method to enhance reasoning performance.

%% file: appendix.tex
\clearpage
\appendix

\input{appendix/baselines_explanation}

\input{appendix/hyper_setting}

\input{appendix/entropy_gradient_inversion_llama}

\input{appendix/entropy_gradient_inversion_rl_sft_stages_llama}

\input{appendix/analyze_correlation}

\input{appendix/different_model_exp}

\input{appendix/full_accuracy_over_steps}

\input{appendix/layerwise_corrpo}

%% file: appendix/baselines_explanation.tex
\section{Appendix}
\subsection{A: Baselines Explanation}

\label{app:baselines_explanation}

We provide a detailed explanation of the baselines compared in Section~\ref{sec:main_exp_math_qwen25_7b} below. All baselines share the same training corpus, rollout protocol, evaluation benchmarks, and backbone models as CorR-PO, and differ only in the reinforcement learning objective or advantage estimation strategy.

\begin{itemize}
    \item \textbf{Base.} The unaligned backbone model (\textit{e.g.}, Qwen2.5-7B-Math, Qwen2.5-14B, Qwen3-4B, Qwen3-1.7B) without any reinforcement learning or reasoning-specific post-training. It reflects the intrinsic reasoning capability of the pretrained model and serves as the lower bound for all RL-based methods, allowing us to isolate the gain contributed purely by the policy-optimization stage.

    \item \textbf{GRPO}~\citep{guo2024deepseekmath}. Group Relative Policy Optimization dispenses with a learned value network and estimates the advantage of each response by standardizing its rule-based reward within a group of $G$ rollouts sampled from the same prompt, \textit{i.e.}, $A_i = (R_i-\text{mean}(R))/\text{std}(R)$. It then applies a token-level importance ratio $r_{i,t}=\pi_\theta(o_{i,t}|q,o_{i,<t})/\pi_{\text{old}}(o_{i,t}|q,o_{i,<t})$ inside a PPO-style min-clip objective together with a KL penalty to the reference policy, and is the de-facto RL algorithm for training large reasoning models.

    \item \textbf{DAPO}~\citep{yu2025dapoopensourcellmreinforcementdapo}. Decoupled Clip and Dynamic Sampling Policy Optimization is a GRPO-family algorithm that decouples the lower and upper clip thresholds via $\text{clip}(r_{i,t},1-\varepsilon_{\text{low}},1+\varepsilon_{\text{high}})$ with $\varepsilon_{\text{low}}<\varepsilon_{\text{high}}$, so that the upper bound is relaxed to allow more aggressive exploration of positive-advantage tokens and mitigate the ``Matthew effect'' of tight clipping on rare good tokens. It additionally adopts dynamic sampling to filter out groups with zero-variance rewards and applies a token-level policy-gradient loss to alleviate length bias, achieving stronger long-chain-of-thought reasoning on math benchmarks.

    \item \textbf{Dr.GRPO}~\citep{liu2025understandingr1zeroliketrainingcritical}. Done-Right GRPO identifies two implicit biases in the vanilla GRPO objective: the response-length normalization $1/|o_i|$ that favors longer outputs and the group-standard-deviation divisor $\text{std}(R)$ that amplifies noise on easy or hard prompts. It removes both normalization terms, reducing the advantage to the simpler form $A_i = R_i-\text{mean}(R)$ while keeping all other PPO machinery identical, which yields an unbiased policy gradient and leads to more stable and token-efficient reasoning training.

    \item \textbf{GSPO}~\citep{zheng2025groupsequencepolicyoptimizationgspo}. Group Sequence Policy Optimization replaces the token-level importance ratio with a sequence-level ratio $s_i=\exp\!\left(\tfrac{1}{|o_i|}\sum_{t=1}^{|o_i|}\log\tfrac{\pi_\theta(o_{i,t}|q,o_{i,<t})}{\pi_{\text{old}}(o_{i,t}|q,o_{i,<t})}\right)$, \textit{i.e.}, the geometric mean of all token ratios in a response, and performs clipping and optimization uniformly at the sequence granularity. This reduces the variance caused by the product of token-level ratios and mitigates the reward-hacking and length-exploitation issues observed in token-level GRPO, delivering state-of-the-art reasoning performance among the open-sourced GRPO variants.
\end{itemize}

%% file: appendix/hyper_setting.tex
\subsection{B: Baselines and LoRA Methods Hyperparameter Setting}

\label{app:baselines_hyper_setting}

To ensure a fair comparison, all baselines in Section~\ref{sec:main_exp_math_qwen25_7b} share the same training corpus, rollout group size ($G=8$), clipping threshold ($\epsilon=0.2$), KL penalty coefficient ($\beta=0.04$), batch size of $16$, maximum generation token length of $32768$, and evaluation protocol using pass@1, pass@16, major@16. 
Method-specific hyperparameters are set following the recommended configurations in the original papers: GRPO~\citep{guo2024deepseekmath} uses a learning rate of $1.0\times 10^{-6}$ with group-standardized advantages; DAPO~\citep{yu2025dapoopensourcellmreinforcementdapo} adopts the decoupled clip thresholds $\epsilon_{\text{low}}=0.2$ and $\epsilon_{\text{high}}=0.28$ with dynamic sampling enabled; Dr.GRPO~\citep{liu2025understandingr1zeroliketrainingcritical} removes the response-length and group-standard-deviation normalization terms while keeping the remaining PPO machinery identical to GRPO; and GSPO~\citep{zheng2025groupsequencepolicyoptimizationgspo} replaces the token-level importance ratio with its sequence-level counterpart and uses $\epsilon=3\times 10^{-4}$ as suggested. For CorR-PO, we use learning rate $3.0\times 10^{-6}$ and correlation regularization coefficient $\lambda_{corr}=0.35$, which we identify as the optimal configuration via the hyperparameter analysis in Table~\ref{tab:hyperparameter}.

For parameter-efficient fine-tuning, we employ LoRA~\citep{hu2021loralowrankadaptationlarge} across all methods with a rank $r=64$, scaling factor $\alpha=128$, and a dropout rate of $0.05$. LoRA adapters are applied to all linear layers of the base model backbone. 

%% file: appendix/entropy_gradient_inversion_llama.tex



%% file: appendix/entropy_gradient_inversion_rl_sft_stages_llama.tex
\subsection{C: Entropy-Gradient Inversion through SFT and GRPO stages on Different Model Architecture}

\label{app:entropy_gradient_inversion_step_llama}

To verify that the Entropy-Gradient Inversion dynamics reported in Section~\ref{sec:training_dynamics} 2.3 are not specific to the Qwen2.5 family, we replicate the three-pipeline tracking experiment on Llama3.1-8B~\citep{grattafiori2024llama} under identical settings: the standard R1 pipeline (8000-step SFT followed by 8000-step GRPO), a pure RL (RL-Zero) pipeline, and a pure SFT pipeline, with the Spearman correlation coefficient between token entropy and gradient nuclear norm evaluated every 200 steps.

\input{multi_plots/entropy_gradient_inversion_step_llama}

\textbf{Inversion emerges rapidly in SFT and is solidified by RL on Llama3.1-8B.} As shown in Figure~\ref{fig:entropy_gradient_inversion_step_llama} (Left), the standard R1 pipeline exhibits a clear phase transition during early SFT, with the correlation rapidly dropping from the near-zero base level to a strongly negative state within the first few hundred steps, and the subsequent GRPO stage further pushes the coefficient toward an even deeper negative value, consistent with the trajectory observed on Qwen2.5-7B.

\textbf{Pure RL and pure SFT reproduce the same divergence pattern.} Figure~\ref{fig:entropy_gradient_inversion_step_llama} (Middle) shows that pure RL without SFT warm-up suffers from unstable oscillation in early steps and converges to a substantially weaker inversion than the full R1 pipeline, while Figure~\ref{fig:entropy_gradient_inversion_step_llama} (Right) confirms that pure SFT drives the correlation steadily downward across 8000 steps. These cross-architecture results verify that SFT is the critical stage for establishing Entropy-Gradient Inversion, and RL further strengthens it into the mature ``slow thinking'' fingerprint.

%% file: multi_plots/entropy_gradient_inversion_step_llama.tex
\begin{figure*}[t]
    \centering
      \centering
    \begin{minipage}{1\textwidth}
      \centering
    \includegraphics[width=0.26\textwidth]{figures/pipeline_legend_cropped.pdf}
    \label{fig:pipeline_legend}
  \end{minipage}
  \vskip\baselineskip 
    \subfigure{
        \includegraphics[width=0.42\linewidth]{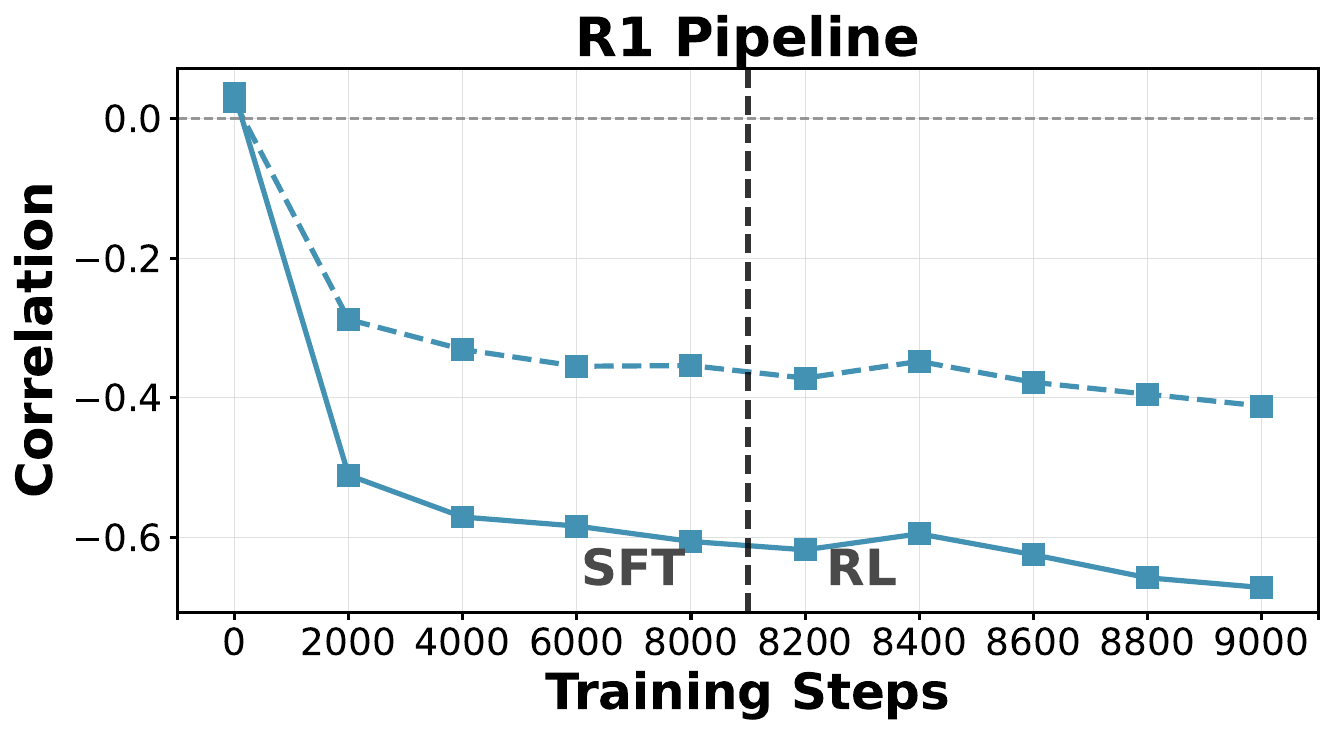}
        \label{fig:r1_pipeline_llama31_8b}
    } 
    \subfigure{
        \includegraphics[width=0.42\linewidth]{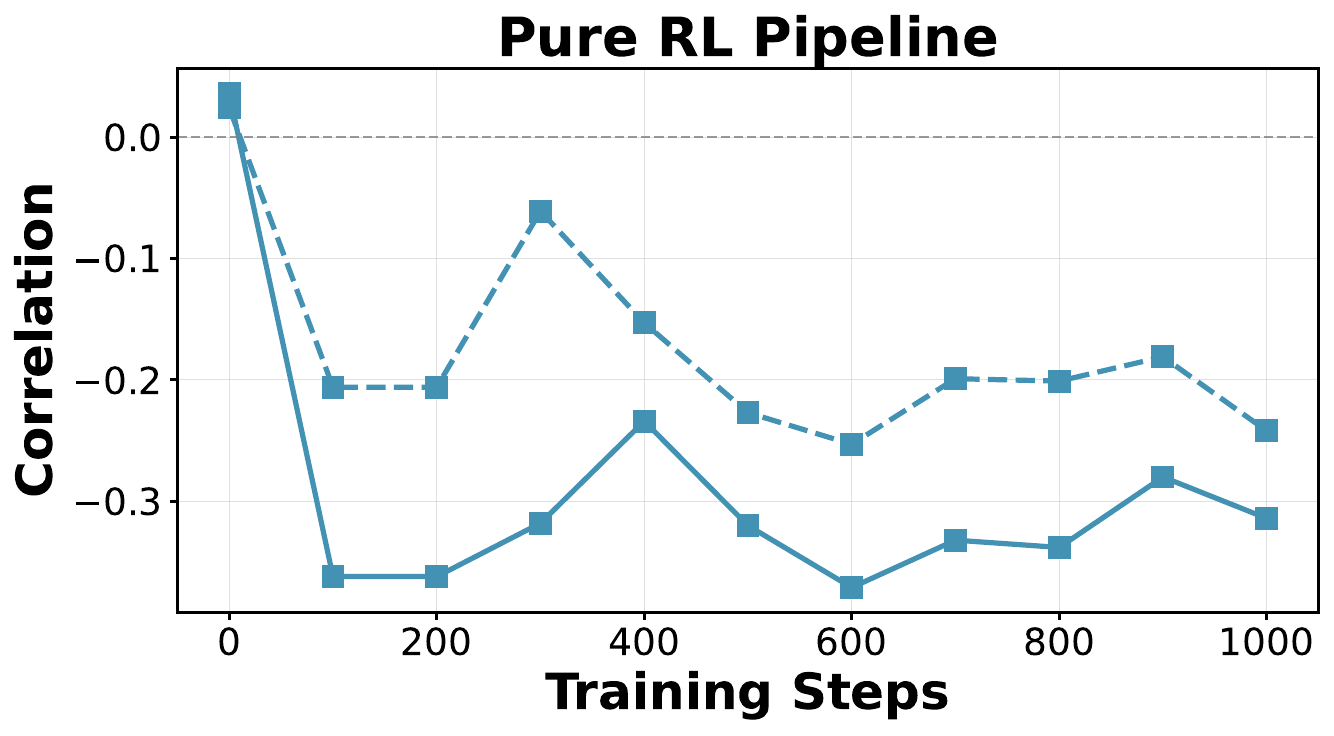}
        \label{fig:pure_rl_pipeline_llama31_8b}
    }
    
    \caption{\textbf{Comparative analysis of correlation variance across three training methodologies using Llama3.1-8B}. \textbf{Left}: The standard DeepSeek-R1 pipeline, consisting of sequential SFT on reasoning data followed by GRPO-based reinforcement learning. \textbf{Right}: A pure RL pipeline where GRPO is applied directly to the base model without an SFT warm-up phase.}
    \label{fig:entropy_gradient_inversion_step_llama}
\end{figure*}

%% file: appendix/analyze_correlation.tex
\subsection{D: Detailed Derivation and Interpretation of Entropy-Gradient Inversion}
\label{app:correlation_analysis}
\subsubsection{Mathematical Derivation of the Entropy-Gradient Relationship.}

To elucidate the internal mechanics of the Entropy-Gradient Inversion, we examine the transformations within the final linear layer of the Large Language Model (LLM). Let $t$ denote the target token, $h \in \mathbb{R}^d$ the hidden state activation vector from the final layer, and $W_t \in \mathbb{R}^d$ the vocabulary weight vector corresponding to token $t$. The logit $z_t$ is computed as:
\begin{equation}
    z_t = W_t^T h
\end{equation}
The probability $p_t$ of the target token is derived via the softmax function:
\begin{equation}
    p_t = \frac{e^{z_t}}{\sum_{k=1}^V e^{z_k}} = \frac{1}{1 + \sum_{k \neq t} e^{z_k - z_t}}
\end{equation}
The Shannon entropy $H$ is defined as $H = -\sum_{k=1}^V p_k \log p_k$. As the model internalizes a reasoning path and reaches a state of high predictive certainty, $H \to 0$ and $p_t \to 1$. For $p_t$ to approach 1, the term $\sum_{k \neq t} e^{z_k - z_t}$ must approach 0, implying $z_t \gg z_k$ for all $k \neq t$. This establishes our first observation: a decrease in entropy $H$ is fundamentally coupled with an increase in the target logit $z_t$.

To analyze the requirements for a large $z_t$, we apply the Cauchy-Schwarz inequality to the inner product:
\begin{equation}
    z_t = |W_t^T h| \le \|W_t\|_2 \|h\|_2
\end{equation}
During inference, the norm of the weight vector $\|W_t\|_2$ is constant. Consequently, achieving a high logit value $z_t$ to minimize entropy necessitates an increase in the norm of the hidden state activation $\|h\|_2$. 

Furthermore, we consider the logit gradient with respect to the weight parameters. For the scalar logit $z_t = \sum_{i=1}^d w_{t,i} h_i$, the partial derivative with respect to a specific weight element $w_{t,k}$ is:
\begin{equation}
    \frac{\partial z_t}{\partial w_{t,k}} = \frac{\partial}{\partial w_{t,k}} \left( \sum_{i=1}^d w_{t,i} h_i \right) = h_k
\end{equation}
Thus, the gradient of the logit with respect to the weight vector $W_t$ is identically the hidden state:
\begin{equation}
    \nabla_{W_t} z_t = h
\end{equation}
Consistent with the definition of the influence metric $I$ used in this study, which measures the intensity of parameter updates via the $L_1$ norm of the gradients, we have:
\begin{equation}
    I \approx \|\nabla_{W_t} z_t\|_1 = \|h\|_1
\end{equation}
This chain of causality reveals the underlying geometric transition:
\begin{equation}
    H \downarrow \xrightarrow{p_t \to 1} z_t \uparrow \xrightarrow{z_t \le \|W\|\|h\|} \|h\| \uparrow \xrightarrow{\nabla_W z_t = h} I \uparrow
\end{equation}
As $H$ decreases, the required activation norm $\|h\|$ increases to sustain the logit magnitude, which in turn elevates the gradient influence $I$. This derivation demonstrates that in reasoning-optimized manifolds, low-entropy states mathematically necessitate high gradient norms, formally justifying the observed negative correlation $\rho_s < 0$.

\subsubsection{Geometric Interpretation of the Inversion.}

The transition from a positive to a negative correlation between entropy and gradient norm suggests a fundamental shift in the model's manifold. In base models, gradients are reactive to local prediction errors. In LRMs, the negative correlation implies that the model's internal geometry is proactively structured; the tokens that drive the logical branching (high entropy) are those where the model is most certain of its reasoning framework, leading to minimal weight perturbation. This inversion provides a concrete, internal geometric metric to distinguish between fast, intuitive text generation and slow, deliberate multi-step reasoning.

%% file: appendix/different_model_exp.tex
\subsection{E: Experiment Results Across Different Model Architectures}
\label{app:different_model_exp}

To further verify that CorR-PO is not tied to a specific model family or parameter scale, we extend our main evaluation to two additional backbones from the Qwen3 series~\citep{yang2025qwen3technicalreport}: \textbf{Qwen3-4B} and \textbf{Qwen3-1.7B}. We keep all training data, benchmarks (AIME24, MATH500, GSM8k), decoding protocols, and evaluation metrics (Pass@1, Pass@16, Major@16) identical to Section~\ref{sec:main_exp_math_qwen25_7b} 4.2, and compare CorR-PO against the unaligned Base model and four RL baselines (GRPO, DAPO, Dr.GRPO, GSPO).

\input{tables/math_performance_experiments/performance_qwen3_4b}

\textbf{CorR-PO matches the strongest baseline on Qwen3-4B.} Table~\ref{tab:performance_qwen3_4b} presents the Pass@1, Pass@16, and Major@16 evaluation results on AIME24, MATH500, and GSM8k benchmarks using Qwen3-4B as the base model. CorR-PO achieves the highest average performance of $79.8$, tying with GRPO while surpassing DAPO, GSPO and Dr.GRPO by $0.3$, $1.5$ and $5.4$ percentage points, and leads on GSM8k Pass@1 ($93.4$) and Major@16 ($95.7$), demonstrating that the proposed correlation regularization remains effective on newer-generation base models.

\input{tables/math_performance_experiments/performance_qwen3_1_7b}

\textbf{CorR-PO remains competitive at the small-scale Qwen3-1.7B regime.} Table~\ref{tab:performance_qwen3_1_7b} shows that CorR-PO reaches an average performance of $67.1$ with Qwen3-1.7B, achieving the second-best result behind GRPO and clearly outperforming DAPO ($66.9$), GSPO ($63.5$) and Dr.GRPO ($62.4$), with the highest Major@16 on MATH500 ($84.2$). These results confirm that CorR-PO's intrinsic correlation-based reward still delivers stable gains in the small-scale regime where advantage-reshaping baselines such as Dr.GRPO and GSPO fail to meaningfully improve over the Base model.

%% file: tables/math_performance_experiments/performance_qwen3_4b.tex
\begin{table*}[ht]
\caption{Main results of CorR-PO and baseline methods on AIME24, MATH500, GSM8k with Qwen3-4B as the base model. All numbers are percentage performances and the best performance among all methods on each dataset is highlighted in \textbf{bold}, while the second-best is marked with \underline{underline}. Average $\uparrow$ column indicate average performance across all benchmarks.}
\label{tab:performance_qwen3_4b}
\renewcommand{\arraystretch}{1.1} 
\resizebox{\textwidth}{!}{\begin{tabular}{l|ccc|ccc|ccc|c}
\toprule
\textbf{Datasets} & \multicolumn{3}{c|}{\textbf{AIME24}} & \multicolumn{3}{c|}{\textbf{MATH500}} & \multicolumn{3}{c|}{\textbf{GSM8k}} & \\
 \midrule
\textbf{Metrics} & Pass@1 & Pass@16 & Major@16 & Pass@1 & Pass@16 & Major@16 & Pass@1 & Pass@16 & Major@16 & Average$\uparrow$ \\
\midrule
\textbf{Base} & 26.7 & 46.7 & 33.3 & 71.8 & 89.8 & 82.0 & 85.5 & 96.3 & 92.9 & 69.4 \\
\textbf{GRPO} & 36.7 & \underline{73.3} & \underline{63.3} & 83.4 & \underline{93.0} & 88.2 & 90.3 & 96.0 & 93.8 & \underline{79.8} \\
\textbf{DAPO} & \underline{53.3} & 60.0 & 50.0 & 86.2 & 93.2 & \underline{90.2} & 91.9 & 96.3 & 94.7 & 79.5 \\
\textbf{Dr.GRPO} & 40.0 & 50.0 & 40.0 & 79.0 & 92.0 & 86.4 & 90.4 & \underline{96.7} & 94.7 & 74.4 \\
\textbf{GSPO} & \underline{46.7} & 56.7 & 46.7 & \underline{86.8} & 93.2 & \underline{90.0} & \underline{92.9} & 96.6 & \underline{95.2} & 78.3 \\
\rowcolor{lightgray!25} \textbf{CorR\textendash PO} & \underline{46.7} & \underline{60.0} & \underline{56.7} & \underline{86.5} & 92.4 & 89.8 & \underline{93.4} & 96.6 & \underline{95.7} & \textbf{79.8} \\
\bottomrule
\end{tabular}}
\end{table*}

%% file: tables/math_performance_experiments/performance_qwen3_1_7b.tex
\begin{table*}[ht]
\centering
\caption{Main results of CorR-PO and baseline methods on AIME24, MATH500, GSM8k with Qwen3-1.7B as the base model. All numbers are percentage performances and the best performance among all methods on each dataset is highlighted in \textbf{bold}, while the second-best is marked with \underline{underline}. Average $\uparrow$ column indicate average performance across all benchmarks.}
\label{tab:performance_qwen3_1_7b}
\renewcommand{\arraystretch}{1.1} 
\resizebox{\textwidth}{!}{\begin{tabular}{l|ccc|ccc|ccc|c}
\toprule
\textbf{Datasets} & \multicolumn{3}{c|}{\textbf{AIME24}} & \multicolumn{3}{c|}{\textbf{MATH500}} & \multicolumn{3}{c|}{\textbf{GSM8k}} & \\
 \midrule
\textbf{Metrics} & Pass@1 & Pass@16 & Major@16 & Pass@1 & Pass@16 & Major@16 & Pass@1 & Pass@16 & Major@16 & Average$\uparrow$ \\
\midrule
\textbf{Base} & 10.0 & 26.7 & 16.7 & 59.8 & 86.8 & 70.2 & 67.6 & 93.5 & 81.8 & 57.0 \\
\textbf{GRPO} & 26.7 & 53.3 & 30.0 & \underline{69.0} & 92.2 & 83.0 & 77.8 & 95.9 & 89.6 & \underline{68.6} \\
\textbf{DAPO} & 20.0 & 43.3 & \underline{26.7} & 70.0 & 93.2 & \underline{83.8} & \underline{79.9} & 95.6 & \underline{89.8} & 66.9 \\
\textbf{Dr.GRPO} & 13.3 & 46.7 & \underline{26.7} & 58.8 & \underline{92.8} & \underline{83.8} & 66.6 & 92.9 & 80.0 & 62.4 \\
\textbf{GSPO} & 13.3 & 40.0 & 23.3 & 62.0 & 91.0 & 80.2 & 76.0 & \underline{96.4} & 89.1 & 63.5 \\
\rowcolor{lightgray!25} \textbf{CorR\textendash PO} & \underline{23.3} & \underline{50.0} & 30.0 & 65.8 & 90.0 & 84.2 & 79.3 & 94.5 & 86.9 & \textbf{67.1} \\
\bottomrule
\end{tabular}}
\end{table*}

%% file: appendix/full_accuracy_over_steps.tex
\subsection{F: CorR-PO Performance Across All Benchmarks through Training Process}
\label{app:full_accuracy_over_steps}

To complement the condensed training-dynamics view in Figure~\ref{fig:step_rl_corr_accuracy}, we report the full per-benchmark and per-metric evaluation of CorR-PO and the GRPO baseline on Qwen2.5-7B-Math at every 100 RL training steps from step 100 to step 1000. Tables~\ref{tab:correlation_rlvr_performance_qwen25_7b_math_step_rl} and~\ref{tab:grpo_performance_qwen25_7b_math_step_rl} list Pass@1, Pass@16 and Major@16 results on AIME24, MATH500 and GSM8k.

\input{tables/step_rl/correlation_rlvr_performance_qwen25_7b_math}

\textbf{CorR-PO achieves consistently higher and more stable performance across training steps.} Table~\ref{tab:correlation_rlvr_performance_qwen25_7b_math_step_rl} shows that the CorR-PO average improves monotonically from $65.8$ at RL-100 to $70.1$ at RL-1000, exceeding $68.0$ at every checkpoint after RL-200 and reaching its peak of $70.1$ at the final step. Notably, AIME24 Pass@1 grows from $13.3$ at RL-100 to $26.7$ at RL-1000 while GSM8k Major@16 remains above $89$ throughout, demonstrating that the correlation regularization continues to deliver gains at later training stages rather than saturating early.

\input{tables/step_rl/grpo_performance_qwen25_7b_math}

\textbf{GRPO saturates early and fluctuates across training.} In contrast, Table~\ref{tab:grpo_performance_qwen25_7b_math_step_rl} indicates that GRPO reaches its best average of $67.0$ already at RL-100 / RL-200 and then oscillates between $63.6$ and $66.8$ without further improvement, with AIME24 Pass@1 fluctuating between $10.0$ and $20.0$ across the remaining 900 steps. Comparing the two tables verifies that the Entropy-Gradient Inversion reward endows CorR-PO with both higher asymptotic accuracy and more stable training dynamics than the GRPO baseline, consistent with the Spearman-correlation trajectory in Figure~\ref{fig:step_rl_corr_accuracy}.

%% file: tables/step_rl/correlation_rlvr_performance_qwen25_7b_math.tex
\begin{table*}[ht]
\centering
\caption{Full evaluation results of CorR-PO on AIME24, MATH500, GSM8k with Qwen2.5-7B-Math as the base model with various training steps. All numbers are percentage performances. Average $\uparrow$ column indicate average performance across all benchmarks.}
\label{tab:correlation_rlvr_performance_qwen25_7b_math_step_rl}
\renewcommand{\arraystretch}{1.1} 
\resizebox{\textwidth}{!}{
\begin{tabular}{l|ccc|ccc|ccc|c}
\toprule
\textbf{Datasets} & \multicolumn{3}{c|}{\textbf{AIME24}} & \multicolumn{3}{c|}{\textbf{MATH500}} & \multicolumn{3}{c|}{\textbf{GSM8k}} & \\
\midrule
\textbf{Metrics} & Pass@1 & Pass@16 & Major@16 & Pass@1 & Pass@16 & Major@16 & Pass@1 & Pass@16 & Major@16 & Average $\uparrow$ \\
\midrule
\textbf{Base} & 10.0 & 40.0 & 20.0 & 60.4 & 90.8 & 79.8 & 82.3 & 97.3 & 90.2 & 63.4 \\
\textbf{RL--100} & 13.3 & 53.3 & 16.7 & 70.0 & 90.6 & 78.8 & 82.6 & 97.2 & 90.1 & 65.8 \\
\textbf{RL--200} & 23.3 & 56.7 & 26.7 & 72.6 & 91.4 & 80.6 & 83.7 & 97.6 & 91.9 & 69.4 \\
\textbf{RL--300} & 23.3 & 56.7 & 26.7 & 71.0 & 90.4 & 81.0 & 83.9 & 97.4 & 92.0 & 69.2 \\
\textbf{RL--400} & 23.3 & 43.3 & 20.0 & 70.8 & 90.8 & 79.2 & 83.5 & 97.2 & 89.8 & 66.4 \\
\textbf{RL--500} & 23.3 & 50.0 & 30.0 & 71.0 & 90.4 & 79.0 & 82.7 & 97.7 & 90.3 & 68.3 \\
\textbf{RL--600} & 23.3 & 53.3 & 33.3 & 72.6 & 89.2 & 80.2 & 82.7 & 97.3 & 91.3 & 69.2 \\
\textbf{RL--700} & 20.0 & 50.0 & 26.7 & 72.7 & 90.4 & 79.6 & 83.5 & 97.9 & 92.4 & 68.1 \\
\textbf{RL--800} & 23.3 & 53.3 & 26.7 & 70.8 & 91.4 & 80.2 & 83.6 & 97.0 & 89.5 & 68.4 \\
\textbf{RL--900} & 26.7 & 53.3 & 33.3 & 73.0 & 92.1 & 78.6 & 83.3 & 97.3 & 89.2 & 69.6 \\
\textbf{RL--1000} & 26.7 & 56.7 & 33.3 & 73.1 & 92.3 & 78.8 & 83.2 & 97.5 & 89.7 & 70.1 \\
\bottomrule
\end{tabular}
}
\end{table*}

%% file: tables/step_rl/grpo_performance_qwen25_7b_math.tex
\begin{table*}[ht]
\centering
\caption{Full evaluation results of GRPO on AIME24, MATH500, GSM8k with Qwen2.5-7B-Math as the base model with various training steps. All numbers are percentage performances. Average $\uparrow$ column indicate average performance across all benchmarks.}
\label{tab:grpo_performance_qwen25_7b_math_step_rl}
\renewcommand{\arraystretch}{1.1}
\resizebox{\textwidth}{!}{
\begin{tabular}{l|ccc|ccc|ccc|c}
\toprule
\textbf{Datasets} & \multicolumn{3}{c|}{\textbf{AIME24}} & \multicolumn{3}{c|}{\textbf{MATH500}} & \multicolumn{3}{c|}{\textbf{GSM8k}} &  \\
\midrule
\textbf{Metrics} & Pass@1 & Pass@16 & Major@16 & Pass@1 & Pass@16 & Major@16 & Pass@1 & Pass@16 & Major@16 & Average $\uparrow$ \\
\midrule
\textbf{Base} & 10.0 & 40.0 & 20.0 & 60.4 & 90.8 & 79.8 & 82.3 & 97.3 & 90.2 & 63.4 \\
\textbf{RL--100} & 20.0 & 40.0 & 30.0 & 71.4 & 91.4 & 81.0 & 84.2 & 96.8 & 88.6 & 67.0 \\
\textbf{RL--200} & 16.7 & 50.0 & 23.3 & 71.4 & 90.6 & 80.2 & 82.8 & 97.5 & 90.6 & 67.0 \\
\textbf{RL--300} & 13.3 & 33.3 & 16.7 & 72.8 & 91.0 & 79.8 & 80.7 & 96.7 & 88.2 & 63.6 \\
\textbf{RL--400} & 10.0 & 46.7 & 26.7 & 76.2 & 88.6 & 75.8 & 84.2 & 97.4 & 90.1 & 66.2 \\
\textbf{RL--500} & 13.3 & 43.3 & 30.0 & 72.8 & 89.8 & 78.6 & 85.5 & 97.3 & 90.4 & 66.8 \\
\textbf{RL--600} & 10.0 & 50.0 & 23.3 & 74.0 & 84.8 & 78.6 & 85.6 & 97.2 & 91.2 & 66.1 \\
\textbf{RL--700} & 13.3 & 43.3 & 20.0 & 74.4 & 92.6 & 81.2 & 85.4 & 96.9 & 91.0 & 66.5 \\
\textbf{RL--800} & 20.0 & 36.7 & 20.0 & 73.6 & 91.4 & 80.4 & 85.8 & 97.3 & 90.9 & 66.2 \\
\textbf{RL--900} & 13.3 & 40.0 & 23.3 & 73.6 & 91.0 & 81.8 & 86.1 & 97.4 & 90.7 & 66.4 \\
\textbf{RL--1000} & 20.0 & 40.0 & 23.3 & 69.0 & 89.8 & 81.0 & 86.0 & 96.7 & 88.1 & 66.0 \\
\bottomrule
\end{tabular}
}
\end{table*}

%% file: appendix/layerwise_corrpo.tex
\subsection{G: CorR-PO Performance Across Different Layers Training}
\label{app:layerwise_corrpo}

To further investigate how the correlation regularization interacts with different transformer depths, we conduct a layer-wise ablation of CorR-PO on Qwen2.5-7B-Math, where the gradient influence $\bar{I}_t$ in the correlation reward is computed from the attention projection gradients of only a single layer $l \in \{1, 2, \dots, 28\}$, while keeping the training data, optimizer, learning rate and $\lambda_{corr}$ identical to the main experiment. The resulting Pass@1 performance on AIME24, MATH500 and GSM8k is summarized in Table~\ref{tab:layerwise_corr_po}.

\input{tables/layerwise_corr_po}

\textbf{CorR-PO consistently improves over the Base model at every individual layer.} As shown in Table~\ref{tab:layerwise_corr_po}, all 28 layer-wise variants lift the average Pass@1 from the Base model's $50.9$ to above $55$, demonstrating that the Entropy-Gradient Inversion signal is a pervasive reasoning fingerprint that is informative at arbitrary depth rather than concentrated in a specific layer.

\textbf{Deeper layers tend to provide stronger supervision for correlation regularization.} The best single-layer configuration, layer 28, reaches an average of $59.6$ with leading AIME24 ($23.3$) and competitive MATH500 ($71.7$) scores, and layers 16, 18, 20 and 23 also exceed $58.0$ on average. This pattern indicates that gradients from deeper transformer blocks more faithfully reflect the structural ``slow thinking'' geometry, yet the full multi-layer aggregation used in the main CorR-PO formulation still outperforms any single-layer variant, justifying the layer-averaged gradient influence design in Section~\ref{subsec:corr_po}3.2.

%% file: tables/layerwise_corr_po.tex
\begin{table*}[ht]
\centering
\caption{Main results of Layer-wise CorR-PO on AIME24, MATH500, GSM8k with Qwen2.5-7B-Math as the base model. All numbers are percentage performances. Average $\uparrow$ column indicate average performance across all benchmarks. The performance of the model is evaluated by Pass@1 of the model.}
\label{tab:layerwise_corr_po}
\renewcommand{\arraystretch}{1.1}
\resizebox{0.6\textwidth}{!}{
\resizebox{\textwidth}{!}{\begin{tabular}{l|ccc|c}
\toprule
\textbf{Layers} & \textbf{AIME24} & \textbf{MATH500} & \textbf{GSM8k} & Average$\uparrow$ \\
\midrule
\textbf{Base Model} & 10.0 & 60.4 & 82.3 & 50.9 \\
\textbf{layer 1} & 13.3 & 70.8 & 83.0 & 55.7 \\
\textbf{layer 2} & 16.7 & 69.8 & 83.9 & 56.8 \\
\textbf{layer 3} & 13.3 & 69.6 & 83.7 & 55.5 \\
\textbf{layer 4} & 16.7 & 71.4 & 82.9 & 57.0 \\
\textbf{layer 5} & 13.3 & 70.0 & 83.7 & 55.7 \\
\textbf{layer 6} & 16.7 & 70.0 & 83.9 & 56.9 \\
\textbf{layer 7} & 13.3 & 70.4 & 84.0 & 55.9 \\
\textbf{layer 8} & 20.0 & 70.2 & 83.4 & 57.9 \\
\textbf{layer 9} & 13.3 & 70.2 & 83.2 & 55.6 \\
\textbf{layer 10} & 20.0 & 71.0 & 84.0 & 58.3 \\
\textbf{layer 11} & 13.3 & 70.4 & 83.3 & 55.7 \\
\textbf{layer 12} & 20.0 & 70.2 & 83.5 & 57.9 \\
\textbf{layer 13} & 13.3 & 70.8 & 83.1 & 55.7 \\
\textbf{layer 14} & 13.3 & 71.6 & 82.7 & 55.9 \\
\textbf{layer 15} & 16.7 & 70.2 & 83.5 & 56.8 \\
\textbf{layer 16} & 20.0 & 71.8 & 83.2 & 58.3 \\
\textbf{layer 17} & 20.0 & 70.0 & 83.3 & 57.8 \\
\textbf{layer 18} & 20.0 & 71.2 & 83.3 & 58.2 \\
\textbf{layer 19} & 20.0 & 69.8 & 82.6 & 57.5 \\
\textbf{layer 20} & 20.0 & 70.6 & 84.2 & 58.3 \\
\textbf{layer 21} & 16.7 & 69.2 & 83.2 & 56.4 \\
\textbf{layer 22} & 16.7 & 68.8 & 83.2 & 56.2 \\
\textbf{layer 23} & 23.3 & 69.8 & 83.5 & 58.9 \\
\textbf{layer 24} & 20.0 & 70.4 & 82.3 & 57.6 \\
\textbf{layer 25} & 20.0 & 71.0 & 83.2 & 58.1 \\
\textbf{layer 26} & 16.7 & 70.4 & 83.7 & 56.9 \\
\textbf{layer 27} & 13.3 & 70.6 & 83.5 & 55.8 \\
\textbf{layer 28} & 23.3 & 71.7 & 83.7 & 59.6 \\
\bottomrule
\end{tabular}}
}
\end{table*}